\RequirePackage[T1]{fontenc}
\documentclass[leqno,twoside,11pt,english]{article}
\usepackage{times,amsmath,amsfonts,amsthm,amssymb,eucal}
\usepackage{graphicx,ifthen}
\usepackage{wrapfig}
\usepackage{latexsym}
\usepackage{xcolor}
\usepackage{mathrsfs}
\usepackage{bm}
\usepackage{subfigure}
\usepackage{rotating}
\usepackage[square,numbers]{natbib}
\usepackage{authblk}
\usepackage{algorithm}
\usepackage{algorithmic}
\usepackage{caption}
\usepackage{url}
\usepackage{babel}
\usepackage{setspace}
\usepackage{booktabs}

\usepackage[hyperfootnotes=false]{hyperref}
\hypersetup{
	colorlinks = true,%
	linkcolor = blue,%
	citecolor = red,%
	linkbordercolor = white,%
	hyperfootnotes = false%
}

\newcommand\blfootnote[1]{%
	\begingroup
	\renewcommand\thefootnote{}\footnote{#1}%
	\addtocounter{footnote}{-1}%
	\endgroup
}

\floatstyle{ruled}
\newfloat{algorithm}{tbp}{loa}
\providecommand{\algorithmname}{Algorithm}
\floatname{algorithm}{\protect\algorithmname}

\newdimen\bigindent
\newdimen\smallindent
\def\quoteindent{\advance\leftskip by\bigindent\advance\rightskip
	by\bigindent}
\newskip\proclaimskipamount
\def\proclaimskip{%
	\par\ifdim\lastskip<\proclaimskipamount
	\removelastskip\vskip\proclaimskipamount\fi}

\def\Demo#1{\par\ifdim\lastskip<\proclaimskipamount
	\removelastskip\proclaimskip\fi
	\sl#1. \hskip\smallindent\rm}

\def\DemoSection#1{\par\ifdim\lastskip<\proclaimskipamount
	\removelastskip\proclaimskip\fi
	#1\hskip\smallindent\rm}
\def\Section#1{\refstepcounter{section}
	\DemoSection{{\bfseries\large\thesection.\hskip\smallindent#1.}}}
\def\Subsection#1{\refstepcounter{subsection}
	\DemoSection{\it\normalsize\thesubsection.\hskip\smallindent#1.}}
\def\Quote{\begin{quotation}\normalfont\small}
	\def\EndQuote{\end{quotation}\rm}

\newtheoremstyle{claim}
{\topsep}
{\topsep}
{}
{}
{\itshape}
{}
{.5em}
{\thmname{#1}\thmnumber{ #2}.\thmnote{ (#3)}}
\theoremstyle{claim}

\newtheorem{theorem}{\smc Theorem}
\newtheorem{proposition}{\smc Proposition}
\newtheorem{lemma}{\smc Lemma}
\newtheorem{corollary}{\smc Corollary}
\newtheorem{remark}{\smc Remark}

\newtheorem{definition}{\smc Definition}

\newtheorem{assumption}{\smc Assumption}

\def\Theorem{\begin{theorem}\sl}
	\def\EndTheorem{\end{theorem}}
\def\Proposition{\begin{proposition}\sl}
	\def\EndProposition{\end{proposition}}
\def\Lemma{\begin{lemma}\sl}
	\def\EndLemma{\end{lemma}}
\def\Corollary{\begin{corollary}\sl}
	\def\EndCorollary{\end{corollary}}
\def\Remark{\begin{remark}\rm}
	\def\EndRemark{\end{remark}}
\def\Definition{\begin{definition}\sl}
	\def\EndDefinition{\end{definition}}

\def\bct{\begin{center}}
	\def\ect{\end{center}}
\def\Array{\begin{eqnarray*}}
	\def\EndArray{\end{eqnarray*}}
\def\Enumerate{\begin{enumerate}}
	\def\EndEnumerate{\end{enumerate}}
\def\Eq{\begin{equation}}
	\def\EndEq{\end{equation}}
\def\EqArray{\begin{eqnarray}}
	\def\EndEqArray{\end{eqnarray}}
\def\Itemize{\begin{itemize}}
	\def\EndItemize{\end{itemize}}
\def\mref#1{(\ref{#1})}
\def\qt#1{\qquad\text{#1}}
\newcommand\addtag{\refstepcounter{equation}\tag{\theequation}}

\def\e{\varepsilon}\def\k{\kappa}
\def\l{\lambda}\def\L{\Lambda}

\def\RSF{\mathscr}

\def\ff{{\RSF F}}
\def\nn{{\mathcal{N}}}

\def\xx{{\RSF X}} 

\def\argmin{\mathop{\rm argmin}}

\def\EE{\mathbb{E}}
\def\PP{\mathbb{P}}
\def\RR{\mathbb{R}}

\def\AGE{\textrm{AGE}}
\def\boost{{\textrm{boost}}}
\def\CENSUS{\textrm{CENSUS}}
\def\ESI{\textrm{ESI}}
\def\NUIS{\textrm{NUISANCE}}
\def\fail{\textrm{Fail}}
\def\ffNF{\tilde{\ff}}
\def\FNF{\tilde{F}}
\def\Fhat{{F}}									
\def\lboosthat{{\hat\lambda}_{\textrm{boost}}}
\def\lhat{{\lambda_n}}
\def\mhat{{\hat m}}
\def\mun{{\mu_n}}								
\def\orlicz#1{{\left\Vert #1 \right\Vert_{\Phi}}}
\def\orliczCond#1#2{{\left\Vert #1 \right\Vert_{\Phi,#2}}}
\def\orliczSqCond#1#2{{\left\Vert #1 \right\Vert^2_{\Phi,#2}}}
\def\Rhat{{R}}									
\def\phihat{{\varphi}}						

\newcommand{\fff}{%
	\mathchoice{\raisebox{0pt}{$\displaystyle\ff$}}
	{\raisebox{0pt}{$\ff$}}
	{\raisebox{-.85pt}{$\scriptstyle\ff$}}
	{\raisebox{-0.4pt}{$\scriptscriptstyle\ff$}}}
\def\aconst{{\alpha_{\!\fff}}}

\font\Caps=cmcsc10
\font\BigCaps=cmcsc10 scaled \magstep 1
\font\HugeCaps=cmcsc9 scaled \magstep 2
\font\BigSlant=cmsl12
 1

\def\smc{\BigCaps}

\begin{document}
\DeclareGraphicsExtensions{.gif,.pdf,.png,.jpg,.tiff}

\thispagestyle{empty}

\begin{center}
	{\bf\Large Boosted nonparametric hazards with time-dependent covariates}\vskip10pt
	{\BigCaps Donald K.K. Lee\footnote[1]{Correspondence: donald.lee@emory.edu. Supported by a hyperplane}, Ningyuan Chen\footnote[2]{Supported by the HKUST start-up fund R9382}, Hemant Ishwaran\footnote[3]{Supported by the NIH grant R01 GM125072}}\vskip5pt
	{\BigSlant Emory University, University of Toronto, University of Miami}\vskip10pt
	\rm 
\end{center}

\vspace*{0.6cm}
\begin{center}
	Preprint of \href{https://doi.org/10.1214/20-AOS2028}{\textit{Annals of Statistics} 49:4:2101-2128 (2021)}
\end{center}
\vspace*{0.3cm}

\Quote
Given functional data from a survival process with time-dependent covariates, we derive a smooth convex representation for its nonparametric log-likelihood functional and obtain its functional gradient. From this we devise a generic gradient boosting procedure for estimating the hazard function nonparametrically.
An illustrative implementation of the procedure using regression trees is described to show how to recover the unknown hazard.  The generic estimator is consistent if the model is correctly specified; alternatively an oracle inequality can be demonstrated for tree-based models. To avoid overfitting, boosting employs several regularization devices. One of them is step-size restriction, but the rationale for this is somewhat mysterious from the viewpoint of consistency. Our work brings some clarity to this issue by revealing that step-size restriction is a mechanism for preventing the curvature of the risk from derailing convergence.
\EndQuote
\vskip1pt

\section{Introduction}\label{sec:Introduction}
Flexible hazard models involving time-dependent covariates are
indispensable tools for studying systems that track covariates over
time.  In medicine, electronic health records systems make it possible
to log patient vitals throughout the day, and these measurements can
be used to build real-time warning systems for adverse outcomes such
as cancer mortality~\citep{adelson2018}. In financial technology,
lenders track obligors' behaviours over time to assess and revise
default rate estimates.  Such models are also used in many other
fields of scientific inquiry since they form the building blocks for
transitions within a Markovian state model.  Indeed, this work was
partly motivated by our study of patient transitions in emergency
department queues and in organ transplant waitlist
queues~\citep{lowsky2013}.  For example, allocation for a donor heart
in the U.S.\ is defined in terms of coarse tiers
\citep{meyer2015future}, and transplant candidates are assigned to
tiers based on their health status at the time of listing.  However, a
patient's condition may change rapidly while awaiting a heart, and
this time-dependent information may be the most predictive
of mortality and not the static covariates collected far in the past.

The main contribution of this paper is to introduce a fully
nonparametric boosting procedure for hazard estimation with
time-dependent covariates.  We describe a generic gradient boosting
procedure for boosting arbitrary base learners for this
setting. Generally speaking, gradient boosting adopts the view of
boosting as an iterative gradient descent algorithm for minimizing a
loss functional over a target function space.  Early work includes
Breiman~\citep{breiman1997BoostingExplained, breiman1999prediction,
	breiman2004population} and Mason et al.~\citep{mason1999functional,mason2000boosting}.  A unified
treatment was provided by Friedman~\cite{friedman}, who coined the term
``gradient boosting'' which is now generally taken to be the modern
interpretation of boosting.

Most of the existing boosting approaches for survival data focus on
time-static covariates and involve boosting the Cox proportional hazards model. Examples
include the popular R-packages {\ttfamily mboost} (B\"uhlmann and Hothorn~\citep{buhlmann2007})
and {\ttfamily gbm} (Ridgeway~\citep{ridgeway}) which apply gradient boosting to the Cox partial likelihood loss.
Related work includes the penalized Cox partial likelihood approach of Binder and Schumacher~\citep{binder2008}.  Other important approaches, but not based on
the Cox model, include $L_2$Boosting~\citep{buhlmann2003} with inverse
probability of censoring weighting
(IPCW)~\citep{hothorn2006}, boosted
transformation models of parametric families \citep{hothorn2019}, and
boosted accelerated failure time models~\citep{huang2006regularized,schmid}.

While there are many boosting methods for dealing with time-static
covariates, the literature is far more sparse for the case of
time-dependent covariates.  In fact, to our knowledge there is no
general nonparametric approach for dealing with this setting.  This is
because in order to implement a fully nonparametric estimator, one has
to contend with the issue of identifying the gradient, which turns out
to be a non-trivial problem due to the functional nature of the data.
This is unlike most standard applications of gradient boosting where
the gradient can easily be identified and calculated.

\Subsection{Time-dependent covariate framework}
To explain why this is so challenging, we start by formally defining the
survival problem with time-dependent covariates.  Our description
follows the framework of Aalen~\cite{aalen1978}.  Let $T$ denote the
potentially unobserved failure time. Conditional on the
history up to time $t-$ the probability of failing at $T \in [t,t+dt)$
equals
\Eq
\lambda(t,X(t))Y(t)dt.
\label{eq:hazard}
\EndEq
Here $\lambda(t,x)$ denotes the unknown hazard function,
$X(t)\in\xx\subseteq\RR^{p}$ is a predictable covariate process, and
$Y(t)\in\{0,1\}$ is a predictable indicator of whether the subject is
at risk at time $t$.\footnote{The filtration of interest is $\sigma
	\{X(s),Y(s),I(T\leq s):s\leq t\}$. If $X(t)$ is only observable when
	$Y(t)=1$, we can set $X(t)=x^c\notin \xx$ whenever $Y(t)=0$.} To
simplify notation, without loss of generality we normalize the units
of time so that $Y(t)=0$ for $t>1$.\footnote{Since the data is always observed up to some finite time, there is no information loss from censoring at that point. For example, if $T'$ is
	the failure time in minutes and the longest duration in the data is
	$\tau'=60$ minutes, the failure time in hours, $T$, is at most
	$\tau=1$ hour. The hazard function on the minute timescale,
	$\lambda_{T'}(t',X(t'))$, can be recovered from the hazard function
	on the hourly timescale, $\lambda_{T}(t,X(t))$, via
	$\lambda_{T'}(t',X(t')) =
	\frac{1}{\tau'}\lambda_{T}(\frac{t'}{\tau'},X(\frac{t'}{\tau'}))$.}
In other words, the subject is not at risk after time $t=1$, so we can
restrict attention to the time interval $(0,1]$.

If failure is observed at $T\in(0,1]$ then the indicator $\Delta=Y(T)$
equals 1; otherwise $\Delta=0$ and we set $T$ to an arbitrary number larger than 1, e.g. $T=\infty$. Throughout we
assume we observe $n$ independent and identically distributed
functional data samples
$\{(X_i(\cdot),Y_i(\cdot),T_i)\}_{i=1}^{n}$. The evolution of
observation $i$'s failure status can then be thought of as a
sequence of coin flips at time increments $t = 0,dt,2dt,\cdots$,
with the probability of ``heads'' at each time point given
by~\eqref{eq:hazard}. Therefore, observation $i$'s contribution to
the likelihood is
$$
\begin{array}{c}
	\{1-\lambda(0,X_i(0))Y_i(0)dt\} \times
	\{1-\lambda(dt,X_i(dt))Y_i(dt)dt\} \times \cdots \times
	\lambda(T_i,X_i(T_i))^{\Delta_i} \\[10pt]
	\xrightarrow[dt\downarrow 0]{}
	e^{-\int_0^1 Y_i(t)\lambda(t,X_i(t))dt}\lambda(T_i,X_i(T_i))^{\Delta_i},
\end{array}
$$
where the limit can be understood as a product integral. Hence, if the
log-hazard function is
$$
F(t,x)=\log\lambda(t,x),
$$
then the (scaled) negative log-likelihood functional is
\Eq
\Rhat_{n}(F)
=\frac{1}{n}\sum_{i=1}^{n}\int_0^{1} Y_i(t)e^{F(t,X_{i}(t))}dt
-\frac{1}{n}\sum_{i=1}^{n}\Delta_{i}F(T_{i},X_{i}(T_{i})) ,
\label{eq:loglik}
\EndEq
which we shall refer to as the likelihood risk. The goal is to
estimate the hazard function $\l(t,x)=e^{F(t,x)}$ nonparametrically by minimizing $\Rhat_n(F)$.

\Subsection{The likelihood does not have a gradient in generic function spaces}
As mentioned, our approach is to boost $F$ using functional gradient
descent.  However, the chief difficulty is that the canonical
representation of the likelihood risk functional does not have a
gradient.  To see this, observe that the directional derivative
of~\mref{eq:loglik} equals
\EqArray
&&\hskip-25pt
\frac{d}{d\theta}\Rhat_n(F+\theta f)\Big|_{\theta=0}\nonumber\\
&& = \frac{1}{n}\sum_{i=1}^{n}\int_0^1 Y_i(t)e^{F(t,X_{i}(t))} f(t,X_{i}(t))dt
- \frac{1}{n}\sum_{i=1}^{n}\Delta_{i}f(T_{i},X_{i}(T_{i})),
\label{eq:directionald}
\EndEqArray
which is the difference of two different inner products $\left\langle
e^F,f\right\rangle _{\dagger}-\left\langle 1,f\right\rangle
_{\ddagger}$ where
\begin{align}
	\langle g,f \rangle_\dagger & = \frac{1}{n} \sum_{i=1}^n \int_0^1 Y_i(t) g(t,X_i(t)) f(t,X_i(t))dt, \nonumber\\
	\langle g,f \rangle_\ddagger & = \frac{1}{n} \sum_{i=1}^n \Delta_i g(T_i,X_i(T_i))f(T_i,X_i(T_i)). \nonumber
\end{align}
Hence,~\eqref{eq:directionald} cannot be expressed as a single inner
product of the form $\langle g_F, f \rangle$ for some function
$g_F(t,x)$.  Were it possible to do so, $g_F$ would then be the gradient function.

In simpler non-functional data settings like regression or
classification, the loss can be written as $L(Y,\FNF(x))$, where
$\FNF$ is the non-functional statistical target and $Y$ is the
outcome, so the gradient is simply $\partial L(Y,\FNF(x))/\partial
\FNF(x)$.  The negative gradient is then approximated using a base
learner $f\in\ffNF$ from a predefined class of functions $\ffNF$ (this
being either parametric; for example linear learners, or
nonparametric; for example tree learners).  Typically, the optimal
base learner $\hat f$ is chosen to minimize the $L^2$-approximation
error and then scaled by a regularization parameter $0<\nu\le 1$ to
obtain the updated estimate of $\FNF$:
$$
\FNF\leftarrow \FNF - \nu \hat f, \hskip15pt \hat
f=\argmin_{f\in\ffNF}\bigg\Vert \frac{\partial L}{\partial
	\FNF}-f\,\bigg\Vert_2.
$$
Importantly, in the simpler non-functional data setting the gradient does
not depend on the space that $\FNF$ belongs to.  By contrast, a key
insight of this paper is that the gradient of $\Rhat_n(F)$ can only
be defined after carefully specifying an appropriate sample-dependent
domain for $\Rhat_n(F)$. The likelihood risk can then be
re-expressed as a smooth convex functional, and an analogous
representation also exists for the population risk. These
representations resolve the difficulty above, allow us to describe and
implement a gradient boosting procedure, and are also crucial to
establishing guarantees for our estimator.

\Subsection{Contributions of the paper}
A key discovery that unlocks the boosted hazard estimator is Proposition~\ref{prop:empR} of Section~\ref{sec:thealgo}. It provides an integral representation
for the likelihood risk from which several results follow, including,
importantly, an explicit representation for the gradient.
Proposition~\ref{prop:empR} relies on defining a suitable space of
log-hazard functions defined on the time-covariate domain
$[0,1]\times\xx$.  Identifying this space is the key insight that
allows us to rescue the likelihood approach and to derive the gradient
needed to implement gradient boosting. Arriving at this framework is
not conceptually trivial, and may explain the absence of boosted
nonparametric hazard estimators until now.

Algorithm~\ref{alg:haz} of Section~\ref{sec:thealgo} describes our estimator. The algorithm minimizes the likelihood risk~\eqref{eq:loglik} over the defined space of log-hazard functions.
In the special case of regression tree learners, expressions for the likelihood risk and its gradient are obtained from Proposition~\ref{prop:empR}, which are then used to describe a tree-based implementation of our estimator in Section~\ref{sec:treeboost}. In Section~\ref{sec:apps} we apply it to a high-dimensional dataset generated from a naturalistic simulation of patient service times in an emergency
department.

Section~\ref{sec:guarantees} establishes the consistency of the
procedure. We show that the hazard estimator is consistent if the
space is correctly specified.  In particular, if the space is the span
of regression trees, then the hazard estimator satisfies an oracle
inequality and recovers $\l$ up to some error tolerance (Propositions~\ref{prop:consistency} and~\ref{prop:oracleineq}).

Another contribution of our work is to clarify the mechanisms used by gradient boosting to avoid overfitting.  Gradient boosting typically applies two types of regularization to invoke slow learning: (i) A small
step-size is used for the update; and (ii) The number of boosting
iterations is capped.  The number of iterations used in our algorithm
is set using the framework of Zhang and Yu~\cite{zhang}, whose work shows how stopping early ensures consistency. On the other hand, the role of step-size restriction is more mysterious. While~\citep{zhang} demonstrates that small step-sizes are needed to prove consistency,
unrestricted greedy step-sizes are already small enough for
classification problems~\citep{telgarsky} and also for commonly used
regression losses (see the Appendix of~\citep{zhang}). We show in
Section \ref{subsec:curvature} that shrinkage acts as a counterweight to the curvature of the risk (see Lemma \ref{lem:emp-converge}). Hence if the curvature is unbounded, as is the case for hazard regression, then the step-sizes may need to be explicitly controlled to ensure convergence.  This important result adds to our understanding of statistical convergence in gradient boosting.  As noted by Biau and Cadre~\cite{biau2017} the literature for this is relatively sparse, which motivated them to propose another regularization mechanism that also prevents overfitting. 

Concluding remarks can be found in Section~\ref{sec:discussion}. Proofs not appearing in the body of the paper can be found in the Appendix.

\Section{The boosted hazard estimator\label{sec:thealgo}}
In this section, we describe our boosted hazard estimator. To provide readers with concrete examples for the ideas introduced here, we will show how the quantities defined in this section specialize in the case of regression trees, which is one of a few possible ways to implement boosting.

We begin by defining in Section~\ref{subsec:dom} an appropriate
sample-dependent domain for the likelihood risk $\Rhat_{n}(F)$.  As
explained, this key insight allows us to re-express the likelihood risk and its population analogue as smooth convex functionals, thereby enabling us to compute their gradients in closed form in
Propositions~\ref{prop:empR} and~\ref{prop:expectedR} of
Section~\ref{subsec:representation}. Following this, the boosting
algorithm is formally stated in Section~\ref{subsec:procedure}.

\Subsection{Specifying a domain for $\Rhat_{n}(F)$}\label{subsec:dom}
We will make use of two identifiability conditions \ref{asm1} and \ref{asm2} to define the domain for $\Rhat_{n}(F)$. Condition \ref{asm1} below is the same as Condition 1(iv)
of Huang and Stone~\cite{huang}.

\begin{assumption}\label{asm1}
	The true hazard function
	$\lambda(t,x)$ is bounded between some interval
	$[\Lambda_L,\Lambda_U]\subset(0,\infty)$ on the time-covariate
	domain $[0,1]\times\xx$.
\end{assumption}

Recall that we defined $X(\cdot)$ and $Y(\cdot)$ to be predictable processes, and so it
can be shown that the integrals and expectations appearing in this
paper are all well defined. Denoting the indicator function as
$I(\cdot)$, define the following population and empirical
sub-probability measures on $[0,1]\times\xx$:
\Array
\mu(B)
&=&\EE \left( \int_0^1 Y(t)\cdot I[\{t,X(t)\}\in B]\,dt \right),\label{eq:mu}\\
\mun(B)
&=&\frac{1}{n}\sum_{i=1}^{n}\int_0^1 Y_i(t)\cdot I[\{t,X_{i}(t)\}\in B]\,dt,
\label{eq:mu_n}
\EndArray
and note that $\EE\mun(B)=\mu(B)$ because the data is
i.i.d.\ by assumption.
Intuitively, $\mun$ measures the denseness of the observed sample
time-covariate paths on $[0,1]\times\xx$. For any integrable $f$,
\EqArray
\int f\,d\mu
&=&\EE \left( \int_{0}^{1} Y(t)\cdot f(t,X(t))\,dt \right),\label{eq:infintegral}\\
\int f\,d\mun
&=&\frac{1}{n}\sum_{i=1}^{n}\int_0^1 Y_i(t) \cdot f(t,X_{i}(t))\,dt.\label{eq:nintegral}
\EndEqArray
This allows us to define the following (random) norms and inner products
\Array
\|f\|_{\mun,1} & = &\int|f|\,d\mun\\
\|f\|_{\mun,2} & = &\left(\int f^{2}\,d\mun\right)^{1/2}\\
\|f\|_{\infty}     & = &\sup\left\{ |f(t,x)|:(t,x)\in[0,1]\times\xx\right\}\\
\left\langle f_1,f_2\right\rangle _{\mun} & = &\int f_1f_2\,d\mun,
\EndArray
and note that
$\|\cdot\|_{\mun,1}\leq\|\cdot\|_{\mun,2}\leq\|\cdot\|_{\infty}$
because $\mun([0,1]\times\xx)\leq1$.

By careful design, $\mun$ allows us to specify a natural domain for
$\Rhat_{n}(F)$. Let $\{\phi_{j}(t,x)\}_{j=1}^{d}$ be a set of bounded
functions $[0,1]\times\xx\mapsto[-1,1]$ that are linearly independent,
in the sense that
$\int_{[0,1]\times\xx}(\sum_{j}c_{j}\phi_{j})^{2}dtdx=0$ if and only
if $c_{1}=\cdots=c_{d}=0$ (when some of the covariates are
discrete-valued, $dx$ should be interpreted as the product of
a counting measure and the Lebesgue measure). The span of the functions is
$$
\ff=\left\{ \sum_{j=1}^{d}c_{j}\phi_{j}:c_{j}\in\RR\right\} .
$$
For example, the span of all regression tree functions that can be defined on $[0,1]\times\xx$ is $\ff=\{ \sum_j c_jI_{B_j}(t,x) : c_j \in \RR \}$,\footnote{It is clear that said span is contained in $\ff$. For the converse, it suffices to show that $\ff$ is also contained in the span of trees of some depth. This is easy to show for trees with $p+1$ splits, because they can generate partitions of the form $(-\infty,t]\times(-\infty,x^{(1)}]\times\cdots\times(-\infty,x^{(p)}]$ in $[0,1]\times\xx$ (Section 3 of \citep{breiman2004population}).} which are linear combinations of indicator functions over disjoint time-covariate cubes indexed\footnote{With a slight abuse of notation, the index $j$ is only considered multi-dimensional when describing the geometry of $B_j$, such as in \eqref{eq:regions}. In all other situations $j$ should be interpreted as a scalar index.} by $j=(j_{0},j_{1},\cdots,j_{p})$:
\begin{equation}
	B_{j}=\left\{ \begin{array}{ccc}
		(t,x) \in [0,1]\times\xx & : & \begin{array}{c}
			t^{(j_{0})}<t\leq t^{(j_{0}+1)}\\
			x^{(1,j_{1})}<x^{(1)}\leq x^{(1,j_{1}+1)}\\
			\vdots\\
			x^{(p,j_{p})}<x^{(p)}\leq x^{(p,j_{p}+1)}
	\end{array}\end{array}\right\}.
	\label{eq:regions}
\end{equation}
\Remark\label{rem:regions}
The regions $B_j$ are formed using all possible split points
$\{x^{(k,j_k)}\}_{j_k}$ for the $k$-th coordinate $x^{(k)}$, with the spacing determined by the precision of the measurements. For example, if weight is measured to the closest kilogram, then the set of all possible split points will be $\{0.5, 1.5, 2.5,\cdots\}$ kilograms. Note that these split points are the finest possible for any realization of weight that is measured to the nearest kilogram. While abstract treatments of trees assume that there is a continuum of split points, in reality they fall on a discrete (but fine) grid that is pre-determined by the precision of the data.
\EndRemark

When $\ff$ is equipped with $\left\langle \cdot,\cdot\right\rangle
_{\mun}$, we obtain the following sample-dependent subspace of
$L^{2}(\mun)$, which is the appropriate domain for
$\Rhat_n(F)$:
$$
(\ff,\left\langle \cdot,\cdot\right\rangle_{\mun}).
\label{eq:hilbertF}
$$
Note that the elements in $(\ff,\left\langle \cdot,\cdot\right\rangle_{\mun})$ are equivalence classes rather than actual functions
that have well defined values at each $(t,x)$. This is a problem
because the likelihood risk~\eqref{eq:loglik} requires evaluating $F(t,x)$
at the points $(T_i,X_i(T_i))$ where $\Delta_i=1$. We resolve this by fixing an orthonormal basis $\{\phihat_{nj}(t,x)\}_{j}$
for $(\ff,\left\langle \cdot,\cdot\right\rangle _{\mun})$, and represent each member of $(\ff,\left\langle \cdot,\cdot\right\rangle _{\mun})$ uniquely in the form
$\sum_{j}c_{j}\phihat_{nj}(t,x)$. For example in the case of regression trees, applying the Gram-Schmidt procedure to $\{\phi_j(t,x) = I_{B_j}(t,x)\}_j$ gives
$$
\{\phihat_{nj}(t,x) \}_j =
\left\{
\frac{I_{B_{j}}(t,x)}{\mun(B_{j})^{1/2}}: \mun(B_{j})>0\right\},
$$
which by design have disjoint support.

The second condition we impose is for $\{\phi_{j}\}_{j=1}^{d}$ to be
linearly independent in $L^{2}(\mu)$, that is
$\|\sum_{j}c_{j}\phi_{j}\|_{\mu,2}^{2} = \sum_{ij}c_{i}\left(\int\phi_{i}\phi_{j}d\mu\right)c_{j}=0$
if and only if $c_{1}=\cdots=c_{d}=0$. Since by construction
$\{\phi_{j}\}_{j=1}^{d}$ are already linearly independent in
$[0,1]\times\xx$, the condition intuitively requires the set of all
possible time-covariate trajectories to be adequately dense in
$[0,1]\times\xx$ to intersect a sufficient amount of the support of
every $\phi_{j}$.  This is weaker than the identifiability conditions 1(ii)-1(iii) in
\citep{huang} which require $X(t)$ to have a positive joint probability
density on $[0,1]\times\xx$.

\begin{assumption}\label{asm2}
	The Gram matrix $\Sigma_{ij}=\int\phi_{i}\phi_{j}d\mu$ is
	positive definite.
\end{assumption}

\Subsection{Integral representations for the likelihood risk}\label{subsec:representation}
Having deduced the appropriate domain for $\Rhat_{n}(F)$, we can now
recast the risk as a smooth convex functional on $(\ff,\left\langle
\cdot,\cdot\right\rangle _{\mun})$. Proposition \ref{prop:empR} below provides closed form expressions for this and its gradient. We note that if the risk is actually of a certain simpler form, it might be possible to estimate its gradient empirically from our risk expression using \citep{mulleryao2010}.

\Proposition\label{prop:empR}
For functions $F(t,x), f(t,x)$ of the form $\sum_j c_j \phihat_{nj}(t,x)$, the likelihood risk~\eqref{eq:loglik} can be written as
\Eq
\Rhat_{n}(F)=\int(e^F-\lhat F)d\mun,
\label{eq:emploglik}
\EndEq
where $\lhat\in(\ff,\left\langle \cdot,\cdot\right\rangle_{\mun})$
is the function
$$
\lhat(t,x)
=\frac{1}{n}\sum_{j}\left\{ \sum_{i=1}^{n}\Delta_{i}\phihat_{nj}(T_{i},X_{i}(T_{i}))\right\} \phihat_{nj}(t,x).
$$
Thus there exists $\rho\in(0,1)$ (depending on $F$ and $f$)
for which the Taylor representation
\begin{equation}
	\Rhat_{n}(F+f)=\Rhat_{n}(F)+
	\left\langle g_{F},f\right\rangle_{\mun}+\frac{1}{2}\int
	e^{F+\rho f}f^{2}d\mun
	\label{eq:emptaylor}
\end{equation}
holds, where the gradient
\begin{equation}
	g_{F}(t,x)
	=\sum_{j}\left\langle e^F,\phihat_{nj}\right\rangle
	_{\mun}\phihat_{nj}(t,x)
	-\lhat(t,x)\label{eq:gradient}
\end{equation}
of $\Rhat_n(F)$ is the projection of $e^F-\lhat$ onto
$(\ff,\left\langle \cdot,\cdot\right\rangle _{\mun})$.
Hence if $g_F=0$ then the infimum of $\Rhat_n(F)$ over
the span of $\{\phihat_{nj}(t,x)\}_{j}$ is uniquely attained
at $F$.
\EndProposition

For regression trees the expressions \eqref{eq:emploglik} and
\eqref{eq:gradient} simplify further because $\ff$ is closed under
pointwise exponentiation, i.e.\ $e^F\in\ff$ for $F\in\ff$. This is
because the $B_j$'s are disjoint so $F = \sum_j c_j I_{B_j}$ and hence
$e^F = \sum_j e^{c_j} I_{B_j}$. Thus
\EqArray
\lhat(t,x) &=&
\sum_{j:\mun(B_j)>0} \frac{\fail_{j}}{n\mun(B_{j})} I_{B_j}(t,x),
\label{eq:mle.hazard}\\
\Rhat_n(F) &=& \sum_{j:{\mun(B_j)>0}}\left(e^{c_{j}}\mun(B_{j})
-\frac{c_{j}\fail_{j}}{n}\right),
\label{eq:treeR}\\
g_F(t,x) &=& \sum_{j:\mun(B_j)>0} \left( e^{c_j} - \frac{\fail_j}{n\mun(B_j)} \right) I_{B_j}(t,x),
\label{eq:treegradient}
\EndEqArray
where
$$
\fail_{j}=\sum_{i}\Delta_{i}I[\{T_{i},X_{i}(T_{i})\}\in B_{j}]
$$
is the number of observed failures in the time-covariate region
$B_{j}$.

\begin{proof}[Proof of Proposition \ref{prop:empR}] Fix a realization of $\{(X_i(\cdot),Y_i(\cdot),T_i)\}_{i=1}^{n}$.
	Using~\eqref{eq:nintegral} we can rewrite~\eqref{eq:loglik} as
	$$
	\Rhat_{n}(F)
	=\int e^{F}d\mun-\frac{1}{n}\sum_{i=1}^{n}\Delta_{i}F(T_{i},X_{i}(T_{i})).
	$$
	We can express $F$ in terms of the basis $\{\phihat_{nk} \}_k$ as
	$F(t,x)=\sum_{k}c_k\phihat_{nk}(t,x)$.  Hence
	\Array
	\int\lhat F\,d\mun
	&=&
	\int \frac{1}{n}\sum_{j}\left\{
	\sum_{i=1}^{n}\Delta_{i}\phihat_{nj}(T_{i},X_{i}(T_{i}))\right\}
	\phihat_{nj}(t,x) F(t,x)\,d\mun\\
	&=&
	\frac{1}{n}\sum_{j}\left\{\sum_{i=1}^{n}\Delta_{i}\phihat_{nj}(T_{i},X_{i}(T_{i}))\right\}
	\int \phihat_{nj}(t,x) F(t,x)\,d\mun\\
	&=&
	\frac{1}{n}\sum_{j}\left\{\sum_{i=1}^{n}\Delta_{i}\phihat_{nj}(T_{i},X_{i}(T_{i}))\right\}
	\int \phihat_{nj}(t,x) \sum_{k}c_k\phihat_{nk}(t,x)\,d\mun\\
	&=&
	\frac{1}{n}\sum_{j}\left\{\sum_{i=1}^{n}\Delta_{i}\phihat_{nj}(T_{i},X_{i}(T_{i}))\right\}
	c_j\\
	&=&
	\frac{1}{n}\sum_{i=1}^{n}\Delta_{i}\sum_{j}c_j\phihat_{nj}(T_{i},X_{i}(T_{i}))\\
	&=&
	\frac{1}{n}\sum_{i=1}^{n}\Delta_{i}F(T_{i},X_{i}(T_{i})),
	\EndArray
	where the fourth equality follows from the orthonormality
	of the basis. This completes the derivation of~\eqref{eq:emploglik}.
	
	By an interchange argument we obtain
	\Array
	\frac{d}{d\theta}\Rhat_{n}(F+\theta f)
	&=&\int(e^{F+\theta f}-\lhat)fd\mun,\\
	\frac{d^{2}}{d\theta^{2}}\Rhat_{n}(F+\theta f)
	&=&\int e^{F+\theta f}f^{2}d\mun,
	\EndArray
	the latter being positive whenever $f\neq0$; i.e.,\ $\Rhat_n(F)$ is
	convex. The Taylor representation~\eqref{eq:emptaylor} then follows
	from noting that $g_F$ is the orthogonal projection of
	$e^{F}-\lhat\in L^{2}(\mun)$ onto $(\ff,\left\langle
	\cdot,\cdot\right\rangle _{\mun})$.
\end{proof}

The expectation of the likelihood risk also has an integral
representation. A special case of the
representation~\eqref{eq:Eloglik} below is proved in Proposition 3.2
of~\citep{huang} for right-censored data only, under assumptions that do not allow for internal covariates. In the statement of the proposition below recall that
$\Lambda_L$ and $\Lambda_U$ are defined in \ref{asm1}.  The constant
$\aconst$ is defined later in~\eqref{eq:alphaconst}.

\Proposition\label{prop:expectedR}
For $F\in\ff\cup\{\log \lambda\}$,
\begin{equation}
	R(F)=\EE\{\Rhat_n(F)\}=\int(e^F-\lambda F)d\mu.\label{eq:Eloglik}
\end{equation}
Furthermore the restriction of $R(F)$ to \textup{$\ff$}
is coercive:
\begin{equation}
	\frac{1}{2}R(F)\geq\frac{\Lambda_{L}}{\aconst}\|F\|_{\infty}
	+\Lambda_U\min\{0,1-\log(2\Lambda_U)\},\label{eq:coercive1}
\end{equation}
and it attains its minimum at a unique point $F^*\in(\ff,\left\langle \cdot,\cdot\right\rangle_\mu)$. If $\ff$
contains the underlying log-hazard function then $F^{*}=\log\lambda$.
\EndProposition

\Remark
Coerciveness~\eqref{eq:coercive1} implies that any $F$ with expected
risk $R(F)$ less than $R(0)\leq1<3$ is uniformly bounded:
\Eq
\|F\|_{\infty}
<\frac{\aconst}{\Lambda_{L}}[3/2+\Lambda_U\max\{0,\log(2\Lambda_U)-1\}]
\leq\aconst\beta_{\Lambda}
\label{eq:coercive2}
\EndEq
where the constant
\begin{equation}
	\beta_{\Lambda}=\frac{3/2+\Lambda_U\max\{0,\log(2\Lambda_U)-1\}}
	{\min\{1,\Lambda_{L}\}}
	\label{eq:betaconst}
\end{equation}
is by design no smaller than 1 in order to simplify subsequent analyses.
\EndRemark

\Subsection{The boosting procedure}\label{subsec:procedure}
In gradient boosting the key idea is to update an iterate in a direction that is approximately aligned to the negative gradient. To model this direction formally, we introduce the concept of an $\e$-gradient.

\Definition\label{def:eps.gradient}
Suppose $g_F\ne 0$. We say that a unit vector $g_F^{\e}\in(\ff,\left\langle
\cdot,\cdot\right\rangle _{\mun})$ is an $\e$-gradient at $F$ if
for some $0<\e\leq1$,
\Eq
\left\langle
\frac{g_F}{\|g_F\|_{\mun,2}},g_F^{\e}
\right\rangle_{\mun}\geq \e.
\label{eq:epsgradient}
\EndEq
Call $-g_F^\e$ a negative $\e$-gradient if $g_F^\e$ is an $\e$-gradient.
\EndDefinition

Our boosting procedure seeks approximations $g_F^{\e}$ that
satisfy~\mref{eq:epsgradient} for some pre-specified alignment value $\e$.
The larger $\e$ is, the closer the alignment is between the negative gradient
and the negative $\e$-gradient, and the greater the risk reduction. In
particular, $-g_F$ is the unique negative 1-gradient with maximal risk reduction.  In practice, however, we find that using a smaller value of $\e$ leads to simpler approximations that prevent overfitting in finite samples. This is consistent with other implementations of boosting: It is well known that the statistical performance of gradient descent generally improves when simpler base learners are used.

Algorithm~\ref{alg:haz} describes the proposed boosting procedure for
estimating $\lambda$. For a given level of alignment $\e$, Line 3
finds an $\e$-gradient $g_{\Fhat_m}^{\e}$ at $\Fhat_m$ satisfying~\eqref{eq:epsgradient.alg} at the $m$-th iteration, and uses its negation for the boosting update in Line 4. If the  $\e$-gradients are tree learners, as is the case with the implementation in Section~\ref{sec:treeboost}, then the trees cannot be grown in the same way as the standard boosting algorithm in Friedman~\cite{friedman}. This is because the standard approach grows all regression trees to a fixed depth, which may or may not ensure $\e$-alignment at each boosting iteration.

To ensure $\e$-alignment, the depth of the trees are not fixed in the implementation in Section \ref{sec:treeboost}. Instead, at each boosting iteration a tree is grown to whatever depth is needed to satisfy~\mref{eq:epsgradient.alg}. This can always be done because the alignment $\e$ is non-decreasing in the number of tree splits, and with enough splits we can recover the gradient $g_{\Fhat_m}$ itself up to $\mun$-almost everywhere.\footnote{Split the tree until each leaf node contains just one of the regions $B_j$ in \eqref{eq:regions} with $\mun(B_j)>0$. Then set the value of the node equal to the value of the gradient function \eqref{eq:treegradient} inside $B_j$.}  As mentioned earlier, we recommend using small values of $\e$, which can be determined in practice using cross-validation. This differs from the standard approach where cross-validation is used to select a common tree depth to use for all boosting iterations.

\vskip10pt
\begin{algorithm}[htpb]
	\centering
	\caption{Boosted nonparametric hazard regression\label{alg:haz}}
	\begin{algorithmic}[1]
		\STATE Initialize $F_0=0$, $m=0$; set $\e\in(0,1]$, and set $\Psi_n$ and $\nu_n$ according to \eqref{eq:stopping} and \eqref{eq:shrinkfactor} respectively
		\WHILE {gradient $g_{F_m}\ne 0$}
		\STATE Compute a weak learner $\e$-gradient $g_{F_m}^{\e}\in(\ff,\left\langle
		\cdot,\cdot\right\rangle_{\mun})$
		satisfying
		\Eq
		\left\langle
		\frac{g_{F_m}}{\|g_{F_m}\|_{\mun,2}},g_{F_m}^{\e}
		\right\rangle_{\mun}\geq \e
		\label{eq:epsgradient.alg}
		\EndEq
		\STATE Compute
		$f\leftarrow F_m-\dfrac{\nu_{n}}{m+1}g_{F_{m}}^{\e}$
		\IF {$\left\Vert f \right\Vert_{\infty} < \Psi_n$}
		\STATE Update the log-hazard estimator:
		$F_{m+1}\leftarrow f$
		\STATE Update $m\leftarrow m+1$
		\ELSE
		\STATE \textbf{break}
		\ENDIF
		\ENDWHILE
		\STATE
		Set $\mhat\leftarrow m$. The estimators for the log-hazard and
		hazard functions are respectively:
		$$
		F_{\mhat}=-\sum_{m=0}^{\mhat-1}\frac{\nu_{n}}{m+1}g_{F_m}^{\e},
		\hskip15pt
		\lboosthat=e^{F_{\mhat}}
		$$
	\end{algorithmic}
\end{algorithm}

In addition to the gradient alignment $\e$, Algorithm~\ref{alg:haz} makes use of two other regularization parameters, $\Psi_n$ and $\nu_n$. The first defines the early stopping criterion (how many boosting iterations to use), while the second controls the step-sizes of the boosting updates. These are two common regularization techniques used in boosting:
\Enumerate
\item
\textit{Early stopping.} The number of boosting iterations
$\mhat$ is controlled by stopping the algorithm before the uniform
norm of the estimator $\|F_{\mhat}\|_{\infty}$ reaches or
exceeds
\Eq
\Psi_{n}=W(n^{1/4})\rightarrow\infty,
\label{eq:stopping}
\EndEq
where $W(y)$ is the branch of the Lambert function that returns the
real root of the equation $ze^z=y$ for $y>0$.
\item
\textit{Step-sizes.}  The step-size $\nu_n \ll 1$ used in gradient
boosting is typically held constant across iterations. While we can also do this in our procedure,\footnote{The term $\nu_n^2 e^{\Psi_n}$ in condition \eqref{eq:shrinkfactor} would need to be replaced by $\mhat\nu_n^2 e^{\Psi_n}$ if a constant step-size is used.} the role of step-size shrinkage becomes more salient if we use $\nu_{n}/(m+1)$ instead as the step-size for the $m$-th iteration in Algorithm~\ref{alg:haz}. This step-size is controlled in two ways. First, it is made to decrease with each iteration according to the Robbins-Monro condition that the sum of the steps diverges while the sum of squared steps converges. Second, the shrinkage factor $\nu_{n}$ is selected to make the step-sizes decay with $n$ at rate
\Eq
\nu_{n}^{2}e^{\Psi_{n}}<1,
\hskip15pt \nu_{n}^{2}e^{\Psi_{n}} \rightarrow 0.
\label{eq:shrinkfactor}
\EndEq
This acts as a counterbalance to $\Rhat_{n}(F)$'s unbounded curvature:
\Eq
\frac{d^{2}}{d\theta^{2}}\Rhat_{n}(F+\theta
f)\Big|_{\theta=0}=\int e^F f^{2}d\mun,
\label{eq:unbounded.curvature}
\EndEq
which is upper bounded by $e^{\Psi_{n}}$ when $\|F\|_{\infty}<\Psi_{n}$
and $\|f\|_{\mun,2}=1$.
\EndEnumerate

\Section{Consistency\label{sec:guarantees}}
Under \ref{asm1} and \ref{asm2}, guarantees for our hazard estimator $\lboosthat$ in Algorithm~\ref{alg:haz} can be derived for two scenarios of interest. The guarantees rely on the regularizations described in Section \ref{subsec:procedure} to avoid overfitting. In the following development, recall from Proposition \ref{prop:expectedR} that $F^*$ is the unique minimizer of $R(F)$, so it satisfies the first order condition
\Eq
\left\langle e^{F^{*}}-\lambda,F\right\rangle _{\mu}=0
\label{eq:L2projection}
\EndEq
for all $F\in\ff$. Recall that the span of all trees is closed under pointwise exponentiation ($e^F\in\ff$), in which case \eqref{eq:L2projection} implies that $\l^* = e^{F^*}$ is the orthogonal projection of $\lambda$ onto $(\ff,\left\langle \cdot,\cdot\right\rangle_{\mu})$.

\Enumerate
\item
{\it Consistency when $\ff$ is correctly specified.}
If the true log-hazard function $\log \lambda$ is in $\ff$,
then Proposition~\ref{prop:expectedR} asserts that
$F^{*}=\log\lambda$. It will be shown in this case
that $\lboosthat$ is consistent:
$$
\left\Vert \lboosthat-\lambda\right\Vert_{\mu,2}^{2} = o_p(1).
\label{eq:informal_consistency}
$$
\item
{\it Oracle inequality for regression trees.} If $\ff$ is closed under pointwise exponentiation, it follows from \eqref{eq:L2projection} that $\lambda^{*}$ is the best $L^{2}(\mu)$-approximation to $\lambda$ among all candidate hazard estimators $\{e^F:F\in\ff\}$. It can then be shown that $\lboosthat$ converges to this best approximation:
$$
\left\Vert \lboosthat -\lambda\right\Vert _{\mu,2}^{2}
=\left\Vert\l^* -\lambda\right\Vert _{\mu,2}^{2}
+ o_p(1).
$$
This oracle result is in the spirit of the type of guarantees available for tree-based boosting in the non-functional data setting. For example, if tree stumps are used for $L_2$-regression, then the regression function estimate will converge to the best approximation to the true regression function in the span of tree stumps \citep{buhlmann2002}. Similar results also exist for boosted classifiers \citep{blanchard2003}.
\EndEnumerate

Propositions~\ref{prop:consistency} and~\ref{prop:oracleineq} below
formalize these guarantees by providing bounds on the error terms above. While sharper bounds may exist, the purpose of this paper is to introduce our generic estimator for the first time and to provide guarantees that apply across different implementations. More refined convergence rates may exist for a specific implementation, just like the analysis in B\"uhlmann and Yu~\cite{buhlmann2003} for $L_2$Boosting when componentwise spline learners are specifically used.

En route to establishing the guarantees, Lemma~\ref{lem:emp-converge} below clarifies the role played by step-size restriction in ensuring convergence of the estimator. As explained in the Introduction, explicit shrinkage is not necessary for classification and regression problems where the risk has bounded curvature. Lemma~\ref{lem:emp-converge} suggests that it may, however, be needed when the risk has unbounded curvature, as is the case with $\Rhat_n(F)$. Seen in this light, shrinkage is really a mechanism for controlling the growth of the risk curvature.

\Subsection{Strategy for establishing guarantees}
The representations for $\Rhat_n(F)$ and its population analogue $R(F)$
from Section \ref{sec:thealgo} are the key ingredients for formalizing the guarantees. We use them to first show that $\Fhat_{\mhat}\in(\ff,\left\langle
\cdot,\cdot\right\rangle_{\mun})$ converges to $F^*\in(\ff,\left\langle
\cdot,\cdot\right\rangle_\mu)$: Applying Taylor's theorem to the representation for $R(F)$ in Proposition~\ref{prop:expectedR} yields
\begin{equation}
	\left\Vert\Fhat_{\mhat}-F^{*}\right\Vert_{\mu,2}^{2}
	\leq
	2\frac{R(\Fhat_{\mhat})-R(F^{*})}{\min_{t,x}(\l^{*}\wedge\lboosthat)}.
	\label{eq:decomp1}
\end{equation}
The problem is thus transformed into one of risk minimization
$R(\Fhat_{\mhat})\rightarrow R(F^{*})$, for which~\cite{zhang}
suggests analyzing separately the terms of the decomposition
\EqArray
0
&\le&
R(\Fhat_{\mhat})-R(F^{*})
\label{eq:decomp2}\\
&\le&
\left|\Rhat_n(\Fhat_{\mhat})-R(\Fhat_{\mhat})\right|
\hskip40pt(\textrm{I})\qquad\text{\Caps complexity argument}
\nonumber\\
&+&
\left|\Rhat_n(F^{*})-R(F^{*})\right|
\hskip40pt(\textrm{II})\qquad\text{\Caps standard argument}
\nonumber\\
&+&
\{\Rhat_n(\Fhat_{\mhat})-\Rhat_n(F^{*})\}.
\hskip25pt(\textrm{III})\qquad\text{\Caps curvature argument}
\nonumber
\EndEqArray
The authors argue that in boosting, the point of limiting the number
of iterations $\mhat$ (enforced by lines 5-10 in Algorithm~\ref{alg:haz}) is to prevent $\Fhat_{\mhat}$ from growing too fast, so that (I) converges to zero as $n\rightarrow\infty$. At the same time, $\mhat$ is allowed to grow with $n$ in a controlled manner so that the empirical risk $\Rhat_{n}(\Fhat_{\mhat})$ in (III) is eventually minimized as $n\rightarrow\infty$. Lemmas~\ref{lem:stat-converge} and~\ref{lem:emp-converge} below show that our procedure achieves both
goals.  Lemma~\ref{lem:stat-converge} makes use of complexity theory
via empirical processes, while Lemma~\ref{lem:emp-converge} deals with
the curvature of the likelihood risk.  The term (II) will be bounded
using standard concentration results.

\Subsection{Bounding (I) using complexity}
To capture the effect of using a simple negative $\e$-gradient~\eqref{eq:epsgradient} as the descent direction, we bound (I) in terms of the complexity of\footnote{
	For technical convenience, $\ff_\e$ has been enlarged from
	$\ff_{\e,\textrm{boost}}$ to include the unit ball.}
\EqArray
&&\ff_\e = \ff_{\e,\boost} \cup \{F\in\ff:\|F\|_{\infty}=1\} \subseteq\ff,
\label{eq:smallF}\\
&&\qquad
\text{where }\ff_{\e,\boost}\
=\left\{\Fhat_m
=-\sum_{k=0}^{m-1}\frac{\nu_{n}}{k+1}g_{\Fhat_k}^{\e}:m = 0,1,\ldots\right\}.
\nonumber
\EndEqArray
Depending on the choice of weak learners for the $\e$-gradients,
$\ff_{\e}$ may be much smaller than $\ff$.  For example, coordinate
descent might only ever select a small subset of basis functions
$\{\phi_{j}\}_{j}$ because of sparsity. As another example if $\l(t,x)$ is additively separable in time and also in each covariate, then regression trees might only ever select simple tree stumps (one tree split).

The measure of complexity we use below comes from empirical process
theory.  Define $\ff_{\e}^{\Psi}=\{F\in\ff_\e:\|F\|_{\infty}<\Psi\}$
for $\Psi>0$ and suppose that $Q$ is a sub-probability measure on
$[0,1]\times\xx$.  Then the $L^{2}(Q)$-ball of radius $\delta>0$
centred at some $F\in L^{2}(Q)$ is
$\{F'\in\ff_{\e}^{\Psi}:\|F'-F\|_{Q,2}<\delta\}$.  The covering
number $\nn(\delta,\ff_{\e}^{\Psi},Q)$ is the minimum number of
such balls needed to cover $\ff_{\e}^{\Psi}$ (Definitions 2.1.5 and
2.2.3 of van der Vaart and Wellner~\cite{vaart}), so $\nn(\delta,\ff_{\e}^{\Psi},Q)=1$ for
$\delta\geq\Psi$. A complexity measure for $\ff_{\e}$ is
\begin{equation}
	J_{\ff_{\e}}=\sup_{\Psi,Q} \left\{
	\int_{0}^{1}\{\log\nn(u\Psi,\ff_{\e}^{\Psi},Q)\}^{1/2}\,du\right\},
	\label{eq:entintegral}
\end{equation}
where the supremum is taken over $\Psi>0$ and over all non-zero
sub-probability measures. As discussed, $J_{\ff_{\e}}$ is never greater than, and potentially much smaller than $J_\ff$, the complexity of $\ff$, which is fixed and finite.

Before stating Lemma~\ref{lem:stat-converge}, we note that the result
also shows an empirical analogue to the norm equivalences
\begin{equation}
	\|F\|_{\mu,1}\leq\|F\|_{\mu,2}\leq\|F\|_{\infty}
	\leq\frac{\aconst}{2}\|F\|_{\mu,1} \text{ for all } F\in\ff
	\label{eq:normequiv}
\end{equation}
exists, where
\begin{equation}
	\aconst
	=2\sup_{F\in\ff:\|F\|_{\infty}=1}\left(\frac{\|F\|_{\infty}}{\|F\|_{\mu,1}}\right)
	=\frac{2}{\displaystyle{\inf_{F\in\ff:\|F\|_{\infty}=1} \|F\|_{\mu,1}}}>1.
	\label{eq:alphaconst}
\end{equation}
The factor of 2 serves to simplify the presentation, and can be
replaced with anything greater than 1.

\vskip6pt
\Lemma\label{lem:stat-converge}
There exists a universal constant
$\kappa$ such that for any $0<\eta<1$, with probability at least
$$
1-4\exp\left\{ -\left(\frac{\eta n^{1/4}}{\kappa\aconst J_{\ff_{\e}}}\right)^{2}\right\}
$$
an empirical analogue to~\eqref{eq:normequiv} holds for all $F\in\ff$:
\begin{equation}
	\|F\|_{\mun,1}\leq\|F\|_{\mun,2}\leq\|F\|_{\infty}\leq\aconst\|F\|_{\mun,1},
	\label{eq:emp-normequiv}
\end{equation}
and for all $F\in\ff_{\e}^{\Psi_{n}}$,
\begin{equation}
	\left|\{\Rhat_{n}(F)-\Rhat_{n}(0)\}-\{R(F)-R(0)\}\right|<\eta.
	\label{eq:stat-converge}
\end{equation}
\EndLemma
\vskip6pt

\Remark\label{rem:dim.equivalence}
The equivalences~\eqref{eq:emp-normequiv} imply that
$\dim(\ff,\left\langle \cdot,\cdot\right\rangle _{\mun})$
equals its upper bound $\dim\ff=d$. That is, if
$\|\sum_{j}c_{j}\phi_{j}\|_{\mun,2}=0$, then
$\|\sum_{j}c_{j}\phi_{j}\|_{\infty}=0$, so $c_{1}=\cdots=c_{d}=0$
because $\{\phi_{j}\}_{j=1}^{d}$ are linearly independent on
$[0,1]\times\xx$.
\EndRemark

\Subsection{Bounding (III) using curvature}\label{subsec:curvature}
We use the representation in Proposition~\ref{prop:empR} to study the
minimization of the empirical risk $\Rhat_n(F)$ by boosting. Standard
results for exact gradient descent like Theorem 2.1.15
of Nesterov~\cite{nesterov} are in terms of the norm of the minimizer, which
may not exist for $\Rhat_{n}(F)$.\footnote{The infimum of $\Rhat_{n}(F)$ is not always attainable: If $f$ is non-positive and vanishes on the set $\{\{T_{i},X_{i}(T_{i})\}:\Delta_{i}=1\}$, then $\Rhat_{n}(F+\theta f)=\int(e^{F+\theta f}-\lhat F)d\mun$ is decreasing in $\theta$ so $f$ is a direction of recession. This is however not an issue for boosting because of early stopping.}  If coordinate descent is used instead, Section 4.1 of~\cite{zhang} can be applied to convex functions whose infimum may not be attainable, but its curvature is
required to be uniformly bounded above. Since the second derivative of
$\Rhat_{n}(F)$ is unbounded~\eqref{eq:unbounded.curvature},
Lemma~\ref{lem:emp-converge} below provides two remedies: (i) Use the
shrinkage decay~\eqref{eq:shrinkfactor} of $\nu_n$ to counterbalance
the curvature; (ii) Use coercivity~\eqref{eq:coercive2} to show that
with increasing probability, $\{\Fhat_m\}_{m=0}^\mhat$ are uniformly
bounded, so the curvatures at those points are also uniformly
bounded. Lemma~\ref{lem:emp-converge} combines both to derive a result
that is simpler than what can be achieved from either one alone. In
doing so, the role played by step-size restriction becomes clear. The
lemma relies in part on adapting the analysis in Lemma 4.1
of~\cite{zhang} for coordinate descent to the case for generic
$\e$-gradients. The conditions required below will be shown to hold
with high probability.

\Lemma\label{lem:emp-converge}
Suppose~\eqref{eq:emp-normequiv} holds and that
$$
\left|\Rhat_n(F^{*})-R(F^{*})\right|<1,
\hskip15pt
\sup_{F\in\ff_\e^{\Psi_{n}}}\left|\Rhat_n(F)-R(F)\right|<1.
$$
Then the largest gap between $F^{*}$ and $\{\Fhat_m\}_{m=0}^\mhat$,
\begin{equation}
	\hat{\gamma}=\max_{m\leq\hat{m}}\|\Fhat_{m}-F^{*}\|_{\infty}\vee1,
	\label{eq:descentgap}
\end{equation}
is bounded by a constant no greater than $2\aconst\beta_{\Lambda}$,
and for $n\geq55$,
\begin{equation}
	\Rhat_{n}(\Fhat_{\mhat})-\Rhat_{n}(F^{*})
	<2e\beta_{\Lambda}\left(\frac{\log
		n}{4n^{1/4}}\right)^{\e/(\aconst\hat{\gamma})}
	+\nu_{n}^{2}e^{\Psi_{n}}.
	\label{eq:emp-converge}
\end{equation}
\EndLemma

\Remark
The last term in~\eqref{eq:emp-converge} suggests that the role of the
step-size shrinkage $\nu_n$ is to keep the curvature of the
risk in check, to prevent it from derailing convergence. Recall
from~\eqref{eq:unbounded.curvature} that $e^{\Psi_{n}}$ describes the
curvature of $\Rhat_{n}(\Fhat_{m})$. Thus our result clarifies the
role of step-size restriction in boosting functional data.
\EndRemark


\Remark
Regardless of whether the risk curvature is bounded or not, smaller
step-sizes always improve the convergence bound. This can be seen from
the parsimonious relationship between $\nu_{n}$
and~\eqref{eq:emp-converge}.  Fixing $n$, pushing the value of
$\nu_{n}$ down towards zero yields the lower limit
$$
2e\beta_{\Lambda}\left(\frac{\log n}{4n^{1/4}}\right)^{\e/(\aconst\hat{\gamma})}.
$$
However, this limit is unattainable as $\nu_n$ must be positive in
order to decrease the risk.
This effect has been observed in practical applications of boosting.
Friedman~\cite{friedman} noted improved performance for gradient boosting with
the use of a small shrinkage factor $\nu$.  At the same time, it was
also noted there was diminishing performance gain as $\nu$ became
very small, and this came at the expense of an increased number of
boosting iterations.  This same
phenomenon has also been observed for
$L_2$Boosting~\citep{buhlmann2003} with componentwise linear learners.
It is known that the solution path for $L_2$Boosting closely matches
that of lasso as $\nu\rightarrow 0$. However, the algorithm exhibits
cycling behaviour for small $\nu$, which greatly increases the number
of iterations and offsets the performance gain in trying to
approximate the lasso (see Ehrlinger and Ishwaran~\cite{ehrlinger2012characterizing}).
\EndRemark

\Subsection{Formal statements of guarantees}
As a reminder, we have defined the following quantities:
{
	\Array
	\lboosthat  & = & \textrm{$e^{\Fhat_{\mhat}}$, the boosted hazard estimator in Algorithm~\ref{alg:haz}} \\
	\l^*        & = & e^{F^*}, \text{ where $F^*$ is the unique minimizer of $R(F)$ in Proposition~\ref{prop:expectedR}}\\
	\Lambda_L,\Lambda_U & = & \text{lower and upper bounds on $\lambda(t,x)$ as defined in \ref{asm1}}\\
	\hat{\gamma}& = & \textrm{maximum gap between $F^{*}$ and
		$\{\Fhat_m\}_{m=0}^\mhat$ defined in~\eqref{eq:descentgap}}\\
	\k          & = & \textrm{a universal constant}\\
	\aconst       & = & \textrm{constant defined in~\eqref{eq:alphaconst}}\\
	\beta_{\L}   & = & \textrm{constant defined in~\eqref{eq:betaconst}}\\
	J_{\ff_\e}    & = & \textrm{complexity measure \eqref{eq:entintegral}, bounded above by } J_{\ff}
	\EndArray}
To simplify the results, we will assume that $n\geq55$ and also set
the shrinkage to satisfy $\nu_{n}^{2}e^{\Psi_{n}}=\log
n/(64n^{1/4})$. Our first guarantee shows that our hazard estimator is consistent if the model is correctly specified.

\Proposition\label{prop:consistency} {\it (Consistency under correct model specification)}. Suppose $\ff$ contains the true log-hazard function $\log\lambda$. Then with probability
$$
1-8\exp\left\{ -\left(\frac{\log n}{\kappa\aconst (\Lambda_L^{-1}\vee\Lambda_U)
	J_{\ff_{\e}}}\right)^{2}\right\}
$$
we have that $\|\lboosthat\|_\infty$ is bounded and
$$
\left\Vert\lboosthat-\l\right\Vert _{\mu,2}^{2}
<13\beta_{\L}\frac{\max_{t,x}(\l\vee\lboosthat)^2}{\min_{t,x}(\l\wedge\lboosthat)}
\left(\frac{\log n}{4n^{1/4}}\right)^{\e/(\aconst\hat{\gamma})}.
$$
Thus $\lboosthat$ is consistent.
\EndProposition


Via the tension between $\e$ and $J_{\ff_\e}$, Proposition~\ref{prop:consistency} captures the trade-off in statistical performance between weak and strong learners in gradient boosting.  The advantage of low complexity (weak learners) is reflected in the increased probability of the $L^2(\mu)$-bound holding, with this probability being maximized when $J_{\ff_\e}\rightarrow0$, which generally occurs as $\e\rightarrow 0$. However, diametrically opposed to this, we find that the $L^2(\mu)$-bound is minimized by $\e\rightarrow 1$, which occurs with the use of stronger learners that are more aligned with the gradient.  This same trade-off is also captured by our second guarantee which establishes an oracle inequality for tree learners.

\Proposition\label{prop:oracleineq} {\it (Oracle inequality for tree learners)}. Suppose $e^F\in\ff$ for $F\in\ff$. Then among $\{e^F:F\in\ff\}$, $\l^*$ is the best $L^2(\mu)$-approximation to $\l$, that is
$$
\l^* = \arg\min_{e^F:F\in\ff}\left\Vert e^F-\lambda\right\Vert _{\mu,2}.
$$
Moreover, $\lboosthat$ converges to this best approximation $\l^*$: With probability
$$
1-8\exp\left\{-\left(\frac{\log n}
{\kappa\aconst(\Lambda_L^{-1}\vee\Lambda_U) J_{\ff_\e}}\right)^{2}\right\}
$$
we have that $\|\lboosthat\|_\infty$ is bounded and
$$
\left\Vert \lboosthat-\lambda\right\Vert _{\mu,2}^{2} <
\rho_\ff^2 +
13\beta_{\Lambda}\frac{\max_{t,x}(\Lambda_U\vee\lboosthat)^{2}}
{\min_{t,x}(\Lambda_L\wedge\lboosthat)}
\left(\frac{\log n}{4n^{1/4}}\right)^{\e/(\aconst\hat{\gamma})},
$$
where $\rho_\ff^2 = \|\l^*-\l\|_{\mu,2}^2$ is the smallest error one can achieve from using functions in $\{e^F:F\in\ff\}$ to approximate $\lambda$.
\EndProposition

For tree learners, $\l^*(t,x)$ is constant over each region $B_j$ in \eqref{eq:regions}, and its value equals the local average of $\l$ over $B_j$,
$$
\l^*(t,x)|_{B_j} = \frac{1}{\mu(B_j)} \int_{B_j}\lambda d\mu.
$$
Hence if the $B_j$'s are small, $\l^*$ should closely approximate $\l$ (recall from Remark \ref{rem:regions} that the size of the $B_j$'s is fixed by the data). To estimate the approximation error $\rho_\ff$ in terms of $B_j$, suppose that $\lambda$ is sufficiently smooth, e.g. H\"older continuous $|\lambda(t,x)-\lambda(t',x')| \precsim \|(t-t',x-x')\|^b$ for some $b>0$. Then since $\inf_{B_j} \lambda \le \l^*|_{B_j} \le \sup_{B_j} \lambda$,
$$
\rho_\ff \leq
\|\lambda^*-\lambda\|_\infty \precsim \max_j(\mbox{diam}B_j)^b.
$$

\Section{A tree-based implementation\label{sec:treeboost}}
Here we describe an implementation of Algorithm~\ref{alg:haz} using
regression trees, whereby the $\e$-gradient $g_{\Fhat_m}^{\e}$ is obtained by growing a tree to satisfy~\mref{eq:epsgradient.alg} for a pre-specified $\e$.

To explain the tree growing process, first observe that the $m$-th step log-hazard estimator is an additive expansion of CART basis functions. Thus it can be written as
\EqArray
\Fhat_m(t,x) &=& \sum_{b=0}^{m-1} \sum_{l=1}^{L_b} \gamma_{b,l} I_{A_{b,l}}(t,x) \nonumber\\
&=&
\sum_j c_{m,j} I_{B_j}(t,x), \label{eq:tree.functionalform}
\EndEqArray
where $A_{b,l}$ is the $l$-th leaf region of the $b$-th tree. Recall from Section \ref{subsec:procedure} that each tree is grown until~\mref{eq:epsgradient.alg} is satisfied, so the number of leaf nodes $L_b$ can vary from tree to tree. The leaf regions are typically large subsets of the time-covariate space $[0,1]\times\xx$ adaptively determined by the tree growing process (to be discussed shortly). Since each leaf region can be further decomposed into the finer disjoint regions $B_j$ in~\mref{eq:regions}, $\Fhat_m(t,x)$ can be rewritten as~\eqref{eq:tree.functionalform}. However, many of these regions will share the same coefficient value, so~\eqref{eq:tree.functionalform} can be written more compactly as
$$
\Fhat_m(t,x) = \sum_j c_{m,j} I_{B_{m,j}'}(t,x),
$$
where $B_{m,j}'$ is the union of contiguous regions whose coefficient equals $c_{m,j}$. This smooths the hazard estimator $\lboosthat(t,x)$ over $[0,1]\times\xx$, thanks to the regularization imposed by limiting the number of trees (early stopping) and also by the use of weak tree learners.  This is unlike the unconstrained hazard MLE $\lhat(t,x)$ defined in~\mref{eq:mle.hazard}, which can take on a different value in each region $B_j$, making it prone to overfit the data.

To construct an $\e$-gradient $g_{\Fhat_m}^{\e}$ with $\e$-alignment to $g_{\Fhat_m}$ defined by~\eqref{eq:treegradient},
$$
g_{\Fhat_m}(t,x) =
\sum_{j:\mun(B_j)>0} \left( e^{c_{m,j}} -
\frac{\fail_j}{n\mun(B_j)} \right) I_{B_j}(t,x),
$$
the tree splits are adaptively chosen to reduce the $L^2(\mun)$-approximation error between $g_{\Fhat_m}^{\e}$ and $g_{\Fhat_m}$. We implement tree splits for both time and covariates. Specifically, suppose we wish to split a leaf region $A \subseteq [0,1]\times\xx$ into left and right daughter subregions $A_1$ and $A_2$, and assign values $\gamma_1$ and $\gamma_2$ to them.  For example, a split on the $k$-th covariate could propose left and right daughters such as
\Eq
A_1=\{(t,x)\in A: x^{(k)}\le s\},\qt{}
A_2=\{(t,x)\in A: x^{(k)}> s\},
\label{covariate.split.point}
\EndEq
or a split on time $t$ could propose regions
\Eq
A_1=\{(t,x)\in A: t\le s\},\qt{}
A_2=\{(t,x)\in A: t> s\}.
\label{time.split.point}
\EndEq
Now note that $g_{\Fhat_m}$ is constant within each region $B_j$. We denote its value by $g_{\Fhat_m}(t_{B_j},x_{B_j})$ where $(t_{B_{j}},x_{B_{j}})$ is the centre of $B_{j}$. Hence the best split of $A$ into $A_1$ and $A_2$ is the one that minimizes
\begin{align*}
	&\hskip-35pt
	\min_{\gamma_1} \int_{A_1} \left\{ g_{\Fhat_m}(t,x)
	- \gamma_1 \right\}^2 d\mun + \min_{\gamma_2} \int_{A_2} \left\{g_{\Fhat_m}(t,x)
	- \gamma_2 \right\}^2 d\mun \\
	&=
	\min_{\gamma_1} \sum_{j:B_j\subseteq A_1} \mun(B_j) \cdot \left\{ g_{\Fhat_m}(t_{B_j},x_{B_j}) - \gamma_1 \right\}^2 \\
	&\qquad+
	\min_{\gamma_2} \sum_{k:B_k\subseteq A_2} \mun(B_k)\cdot \left\{ g_{\Fhat_m}(t_{B_k},x_{B_k}) - \gamma_2 \right\}^2 \\
	&=
	\min_{\gamma_1} \sum_{\substack{j:z_j\in A_1,\\w_j>0}} w_j\cdot (\tilde y_j - \gamma_1)^2
	+ \min_{\gamma_2} \sum_{\substack{k:z_k\in A_2,\\w_k>0}} w_k\cdot (\tilde y_k - \gamma_2)^2, \addtag\label{eq:weightedLS}
\end{align*}
where
$$
\tilde y_j = g_{\Fhat_m}(t_{B_{j}},x_{B_{j}})
= e^{c_{m,j}} - \frac{\fail_j}{n\mun(B_j)}
$$
represents the $j$-th pseudo-response, $z_j = (t_{B_{j}},x_{B_{j}})$
its covariate and $w_j = \mun(B_j)$ its weight. Thus the splits use a weighted least squares criterion, which can be efficiently computed as usual.

We split the tree until~\mref{eq:epsgradient.alg} is satisfied, resulting in $L_m$ leaf nodes ($L_m-1$ splits).  As discussed in Section~\ref{subsec:procedure}, we can always find a deep enough tree that is an $\e$-gradient because with enough splits we can recover the gradient $g_{\Fhat_m}$ itself. Recall also that a small value of $\e$ performs best in practice, and this can be chosen by cross-validating on a set of small-sized candidates: For each one we implement Algorithm~\ref{alg:haz}, and we select the one that minimizes the cross-validated risk $\Rhat_n(F)$ defined in~\eqref{eq:treeR}. By contrast, the standard boosting algorithm~\citep{friedman} uses cross-validation to select a common number of splits to use for all trees, which does not ensure that each tree is an $\e$-gradient.

Regarding the possible split points for the covariates~\mref{covariate.split.point}, note that the $k$-th covariate $x^{(k)} = x^{(k)}(t)$ is a time series that is sampled periodically. This yields a set of unique values equal to the union of all of the sampled values for the $n$ observations. In direct analogy to non-functional data boosting, we place candidate split points in-between the sorted values in this set.  In other words, splits for covariates only occur at values corresponding to the observed data just as in non-functional boosting.

The resolution for the grid of candidate time
splits~\mref{time.split.point} is set equal to the temporal
resolution. For example, the covariate trajectories in the simulation
in Section~\ref{sec:apps} are piecewise constant and may change every
0.002 days. Placing the candidate split points at $\{0.002,0.004,\ldots\}$ days simplifies the exact computation of $\mun(B_j)$
because every covariate trajectory is constant between these points.
Again, notice that the splits for time only occur at values informed by
the observed data.

Putting it together, the setup above leverages our insight in
\eqref{eq:weightedLS} by transforming the survival functional data
into the data values $\{w_j, \tilde y_j, z_j\}_{j:w_j>0}$, which
enables the implementation to proceed like standard gradient boosting
for non-functional data. Only the pseudo-response $\tilde y_j$ in
$\{w_j, \tilde y_j, z_j\}$ needs to be updated at each boosting
iteration, while the other two do not change. In terms of storage it
costs $\mathcal{O}(np|\mathcal{T}|)$ to store $\{w_j, \tilde y_j,
z_j\}_{j:w_j>0}$, where $|\mathcal{T}|$ is the cardinality of the set
of candidate time splits.\footnote{Each $\{w_j, \tilde y_j, z_j\}$ is
	of dimension $p+3$ and the number of time-covariate regions $B_j$
	with $w_j>0$ is at most $n(|\mathcal{T}|+1)$. To show the latter,
	observe that $B_j$ will only have $w_j=\mun(B_j)>0$ if it is
	traversed by at least one sample covariate trajectory. Then note
	that each of the $n$ sample covariate trajectories can traverse at
	most $|\mathcal{T}|+1$ unique regions.} Computationally, choosing a
new tree split requires testing $\mathcal{O}(np|\mathcal{T}|)$
candidate splits.\footnote{A sample covariate trajectory can have at
	most $|\mathcal{T}|$ unique observed values for the $k$-th covariate
	$x^{(k)}$, so there are at most $n|\mathcal{T}|$ candidate splits
	for $x^{(k)}$. Thus there are $\mathcal{O}(np|\mathcal{T}|)$
	candidate splits for $p$ covariates. The number of candidate splits
	on time is obviously $|\mathcal{T}|$.} The space and time
complexities of the implementation are reasonable given that they are
$\mathcal{O}(np)$ for non-functional data boosting: In the functional
data setting, each sample can have up to $|\mathcal{T}|$ observations,
so $n$ functional data samples is akin to
$\mathcal{O}(n|\mathcal{T}|)$ samples in a non-functional data
setting.

\Section{Numerical experiment\label{sec:apps}}
We now apply the boosting procedure of Section~\ref{sec:treeboost} to a
high-dimensional dataset generated from a naturalistic
simulation. This allows us to compare the performance of our estimator
to existing boosting methods. The simulation is of patient service
times in an emergency department (ED), and the hazard function of
interest is patient service rate in the ED. The study of patient
transitions in an ED queue is an important one in healthcare
operations, because without a high resolution model of patient flow
dynamics, the ED may be suboptimally utilized which would deny patients
of timely critical care.

\Subsection{Service rate} The service rate model used in the simulation is based upon a service time dataset from the ED of an academic hospital in the United States. The dataset contains information on 86,983 treatment encounters from 2014 to early 2015. Recorded for each encounter was: Age, gender, Emergency Severity Index (ESI)\footnote{Level 1 is the most severe
	(e.g., cardiac arrest) and level 5 is the least (e.g., rash). We
	removed level 1 patients from the dataset because they were treated
	in a separate trauma bay.}, time of day when treatment in the ED
ward began, day of week of ED visit, and ward census. The last one
represents the total number of occupied beds in the ED ward, which
varies over the course of the patient's stay. Hence it is a
time-dependent variable. Lastly, we also have the duration of the
patient's stay (service time).

The service rate function is developed from the data in the following
way. First, we apply our nonparametric estimator to the data to
perform exploratory analysis.  We find that:
\Enumerate
\item The key variables affecting the service rate (based on relative variable importance \citep{friedman}) are ESI, age, and ward census. In addition, two of the most pronounced interaction terms identified by the tree splits are $(\AGE\ge34, \ESI=5)$ and $(\AGE\ge34, \ESI\le4)$.
\item Holding all the variables fixed, the shapes of the estimated service rate function resemble the hazard functions of log-normal distributions. This agrees with the queuing literature that find log-normality to be a reasonable parametric fit for service durations.
\EndEnumerate
Guided by these findings, we specify the service rate
$\lambda(t,X(t))$ for the simulation as a log-normal accelerated
failure time (AFT) model, and estimate its parameters from data.
This yields the service rate
\begin{equation}\label{eq:simhaz}
	\lambda(t,x) = \theta(x) \cdot \frac{\phi_l(\theta(x)t;m,\sigma)}{1-\Phi_l(\theta(x)t;m,\sigma)},
\end{equation}
where $\phi_l(\cdot;m,\sigma)$ and $\Phi_l(\cdot;m,\sigma)$ are the
PDF and CDF of the log-normal distribution with log-mean $m=-1.8$ and
log-standard deviation $\sigma=0.74$. The function $\theta(x)$ captures
the dependence of the service rate on the covariates:
\begin{equation}\label{eq:simspec}
	\begin{aligned}
		\log\theta(X(t)) &= -0.0071\cdot \AGE + 0.022\cdot \ESI - \min \left\{ a\cdot \frac{\CENSUS_t}{70},2 \right\} \\
		&\qquad + 0.10\cdot I(\AGE\ge 34,\ESI=5) - 0.10\cdot I(\AGE\ge 34,\ESI\le 4) \\
		&\qquad + 0\cdot\NUIS_1 + \cdots + 0\cdot\NUIS_{43}.
	\end{aligned}
\end{equation}
The specification for $\theta(X(t))$ above is a slight modification of
the original estimate, with the free parameter $a$ allowing us to
study the effect of time-dependent covariates on hazard
estimation. When $a=0$, the service rate does not depend on
time-varying covariates, but as $a$ increases, the dependency becomes
more and more significant. In the data, the ward census never exceeds
70, so we set the capacity of the simulated ED to 70 as well. The
$\min$ operator caps the impact that census can have on the simulated
service rate as $a$ grows. The irrelevant covariates $\NUIS_1,\cdots,\NUIS_{43}$ are added to the data in order to assess how boosting performs in high dimensions. We explicitly include them in \eqref{eq:simspec} to remind ourselves that the simulated data is high-dimensional. Forty of the irrelevant variables are generated synthetically as described in the next subsection, while the rest are variables from the original dataset not used in the simulation.

\Subsection{Simulation model}
Using~\eqref{eq:simhaz} and \eqref{eq:simspec}, we simulate a naturalistic
dataset of 10,000 patient visit histories. The value of $a$ will be
varied from 0 to 3 in order to study the impact of time-dependent
covariates on hazard estimation. Each patient is associated with a
46-dimensional covariate vector consisting of:
\Itemize
\item The time-varying ward census. The initial value is sampled from its marginal empirical distribution in the original dataset. To simulate its trajectory over a patient's stay, for every timestep advance of 0.002 days ($\approx$3 minutes), a Bernoulli(0.02) random variable is generated. If it is one, then the census is incremented by a normal random variable with zero mean and standard deviation 10. The result is truncated if it lies outside the range $[1,70]$, the upper end being the capacity of the ED.
\item The other five time-static covariates in the original dataset. These are sampled from their marginal empirical distributions in the original dataset. Two of the variables (age and ESI) influence the service rate, while the other three are irrelevant.
\item An additional forty time-static covariates that do not affect the service rate (irrelevant covariates). Their values are drawn uniformly from [0,1].
\EndItemize

We also generate independent censoring times (rounded to the nearest 0.002 days) for each visit from an exponential distribution. For each simulation, the rate of the exponential distribution is set to achieve an approximate target of 25\% censoring.

\Subsection{Comparison benchmarks}
When the covariates are static in time, a few software packages are available for performing hazard estimation with tree ensembles. Given that the data is simulated from a log-normal hazard, we compare our nonparametric method to two correctly specified parametric estimators:
\Enumerate
\item The {\ttfamily blackboost} estimator in the R package {\ttfamily mboost}~\citep{buhlmann2007} provides a tree boosting procedure for fitting the log-normal hazard function. In order to apply this to the simulated data, we make ward census a time-static covariate by fixing it at its initial value.
\item Transformation forests~\citep{hothorn2017} in the R package {\ttfamily trtf} can also fit log-normal hazards. Moreover, it allows for left-truncated and right-censored data. Since the ward census variable is simulated to be piecewise constant over time, we can treat each segment as a left-truncated and right-censored observation. Thus for this simulation, transformation forests are able to handle time-dependent covariates with time-static effects. This falls in between the static covariate/static effect {\ttfamily blackboost} estimator and our fully nonparametric one.
\EndEnumerate

Since the service rate model used in the simulations is in fact log-normal, the benchmark methods above enjoy a significant advantage over our nonparametric one, which is not privy to the true distribution. In fact, when $a=0$ the log-normal hazard \eqref{eq:simhaz} depends only on time-static covariates, so the benchmarks should outperform our nonparametric estimator. However, as $a$ grows, we would expect a reversal in relative performance.

To compare the performances of the estimators, we use Monte Carlo integration to evaluate the relative mean squared error
$$
\text{\%MSE} = \frac{\mathbb{E}_X\left[\int_0^1 \{\lambda(t,X)-\hat{\lambda}(t,X)\}^{2}dt\right]}
{\mathbb{E}_X \left[\int_0^1 \lambda(t,X)^{2}dt\right]}.
$$
The Monte Carlo integrations are conducted using an independent test set of 10,000 uncensored patient visit histories. For the test set, ward census is held fixed over time at the initial value, and we use the grid $\{0,0.02,0.04,\cdots,1\}$ for the time integral. The nominator above is then estimated by the average of $\{\lambda(t,x)-\hat{\lambda}(t,x)\}^2$ evaluated at the 51$\times$10,000 points of $(t,x)$. The denominator is estimated in the same manner.

\Subsection{Results}
For the implementation of our estimator in Section \ref{sec:treeboost}, the value of $\e$ and the number of trees $\hat m$ are jointly determined using ten-fold cross validation.  The candidate values we tried for $\e$ are $\{0.003, 0.004, 0.005, 0.006, 0.007\}$, and we limit $\hat m$ to no more than 1,000 trees.  A wider range of values can be of course be explored for better performance  (at the cost of more computations). As comparison, we run an ad-hoc version of our algorithm in which all trees use the same number of splits, as is the case in standard boosting~\citep{friedman}.  This approach does not explicitly ensure that the trees will be $\e$-gradients for a pre-specified $\e$. The number of splits and the number of trees used in the ad-hoc method are jointly determined using ten-fold cross-validation.

In order to speed up convergence at the $m$-th iteration for both approaches, instead of using the step-size $\nu_{n}/(m+1)$ of Algorithm~\ref{alg:haz}, we performed line-search within the interval $(0,\nu_n/(m+1)]$.  While Lemma~\ref{lem:emp-converge} shows that a smaller shrinkage $\nu_n$ is always better, this comes at the expense of a larger $\hat m$ and hence computation time. For simplicity we set $\nu_n=1$ for all the experiments here.

For fitting the {\ttfamily blackboost} estimator, we use the default setting of {\ttfamily nu} $=0.1$ for the step-size taken at each iteration. The other hyperparameters, {\ttfamily mstop} (the number of trees) and {\ttfamily maxdepth} (maximum depth of trees), are chosen to directly minimize the relative MSE on the test set. This of course gives the {\ttfamily blackboost} estimator an unfair advantage over our estimator, which is on top of the fact that it is based on the same distribution as the true model. Transformation forest (using also the true distribution) is fit using code kindly provided by Professor T. Hothorn.\footnote{In the code 100 trees are used in the forest, which takes about 700 megabytes to store the fitted object when applied to our simulated data.}

\textit{Variable selection}.  The relative importance of
variables~\citep{friedman} for our estimator are given in
Table~\ref{tab:rel_imp} for all four cases $a=0,1,2,3$. The four
factors that influence the service rate \eqref{eq:simspec} are
explicitly listed, while the irrelevant covariates are grouped
together in the last column. When $a=0$, the service rate does not
depend on census, and we see that the importance of census and the
other irrelevant covariates are at least an order of magnitude smaller
than the relevant ones. As $a$ increases, census becomes more and more
important as correctly reflected in the table. Across all the cases
the importance of the relevant covariates are at least an order of
magnitude larger than the others, suggesting that our estimator is
able to pick out the influential covariates and largely avoid the
irrelevant ones.
\begin{table}[]
	\caption{\sl Relative importance of variables in the boosted nonparametric estimator. The numbers are scaled so that the largest value in each row is 1.}
	\centering
	\begin{tabular}{ccccccc}
		\toprule
		$a$ & Time & Age & ESI & Census & All other variables \\
		\midrule
		0 & 1 & 0.21 & 0.025 & 0.0011 & <0.0010\\
		1 & 1 & 0.22 & 0.013 & 0.46 & <0.0003\\
		2 & 0.34 & 0.064 & 0.0020 & 1 & 0\\
		3 & 0.11 & 0.011 & <0.0001 & 1 & 0\\
		\bottomrule
	\end{tabular}
	\label{tab:rel_imp}
\end{table}

\begin{table}[]
	\caption{\sl Comparative performances (\%MSE) as the service rate~\mref{eq:simspec} becomes increasingly dependent on the time-varying ward census variable (by increasing $a$).}
	\centering
	\begin{tabular}{ccccccc}
		\toprule
		& {\ttfamily blackboost} & {\it Transformation forest} & {\it Boosted hazards} & {\it Ad-hoc}
		\\
		$a$ & (set to true log- & (set to true log- & {\it ($\varepsilon$
			fixed for} & {\it (\# splits fixed}
		\\
		& normal distribution) & normal distribution) & {\it all iterations)} & {\it for all iterations)}
		\\
		\midrule
		0 & 5.0\% & 5.0\% & 7.8\% & 7.1\% \\
		1 & 17\% & 6.1\% & 4.5\%  & 8.1\% \\
		2 & 46\%  & 9.7\% & 5.4\% & 7.0\% \\
		3 & 67\%  & 18\% & 7.2\% & 7.4\% \\
		\bottomrule
	\end{tabular}
	\label{tab:mse1}
\end{table}

\textit{Presence of time-dependent covariates}. Table~\ref{tab:mse1} presents the relative MSEs for the estimators as the service rate function~\eqref{eq:simspec} becomes increasingly
dependent on the time-varying census variable. When $a=0$ the service rate depends only on time-static covariates, so as expected, the parametric log-normal benchmarks perform the best when applied to data simulated from a log-normal AFT model.

However, as $a$ increases, the service rate becomes increasingly
dependent on census. The corresponding performances of both benchmarks deteriorate dramatically, and is handily outperformed by the proposed estimator. We note that the inclusion of just one time-dependent covariate is enough to degrade the
performances of the benchmarks, despite the fact that they have the
exact same parametric form as the true model.

Finally we find comparable performance among the ad-hoc boosted
estimator and our proposed one, although a slight edge goes to the latter especially in the more difficult simulations with larger $a$. The results here demonstrate that there is a place in the survival
boosting literature for fully nonparametric methods like this one that
can flexibly handle time-dependent covariates.

\Section{Discussion\label{sec:discussion}}
Our estimator can also potentially be used to evaluate the
goodness-of-fit of simpler parametric hazard models. Since our
approach is likelihood-based, future work might examine whether model
selection frameworks like those in~\citep{vuong} can be extended to
cover likelihood functionals. For this,~\citep{buhlmann2007} provides
some guidance for determining the effective degrees of freedom for the
boosting estimator. The ideas in~\citep{zou2007} may also be germane.

The implementation presented in Section \ref{sec:treeboost} is one of many possible ways to implement our estimator. We defer the design of a more refined implementation to future research, along with open-source code.

{\vskip10pt\bf\large Acknowledgements.} The review team provided many insightful comments that significantly improved our paper. We are grateful to Brian Clarke, Jack Hall, Sahand Negahban, and Hongyu Zhao for helpful discussions. Special thanks to Trevor Hastie for early formative discussions. The dataset used in Section \ref{sec:apps} was kindly provided by Dr. Kito Lord.

\vskip40pt
\centerline{\HugeCaps APPENDIX: PROOFS}

\subsection*{Proof of Proposition \ref{prop:expectedR}}

\begin{proof}
	Writing
	$$
	R(F)=\EE\left( \int_{0}^{1}Y(t)\cdot e^{F(t,X(t))}\,dt-\Delta F(T,X(T)) \right),
	$$
	we can apply~\eqref{eq:infintegral} to establish the first part of the
	integral in~\eqref{eq:Eloglik} when $F\in\ff\cup\{\log\lambda\}$.  To
	complete the representation, it suffices to show that the point
	process
	$$
	M(B)=\Delta\cdot I[\{T,X(T)\}\in B]
	$$
	has mean $\int_{B}\lambda d\mu$, and then apply Campbell's formula. To
	this end, write $N(t)=I(T\leq t)$ and consider the filtration $\sigma
	\{X(s),Y(s),N(s):s\leq t\}$. Then $N(t)$ has the Doob-Meyer form
	$dN(t)=\lambda(t,X(t))Y(t)dt+dM(t)$ where $M(t)$ is a
	martingale. Hence
	\Array
	\EE\{ M(B) \}&=&\EE\left( \int_{0}^{1}I[\{t,X(t)\}\in B]\,dN(t)\right)\\
	&=&\EE \left( \int_{0}^{1} Y(t)\cdot I[\{t,X(t)\}\in B]\cdot \lambda(t,X(t))\,dt \right)\\
	&&+ \EE\left(\int_{0}^{1}I[\{t,X(t)\}\in B]\,dM(t)\right)\\
	&=&\int_{B}\lambda \,d\mu
	+\EE\left(\int_{0}^{1}I[\{t,X(t)\}\in B]\,dM(t)\right),
	\EndArray
	where the last equality follows from \eqref{eq:infintegral}. Since
	$I[\{t,X(t)\}\in B]$ is predictable because $X(t)$ is, the desired
	result follows if the stochastic integral $\int_{0}^{1}I[\{t,X(t)\}\in B]dM(t)$ is a martingale. By Section 2 of Aalen~\cite{aalen1978}, this is true if $M(t)$ is square-integrable. In fact, $M(t)=N(t)-\int_{0}^{t}\lambda(t,X(t))dt$ is bounded because $\lambda(t,x)$ is bounded above by \ref{asm1}. This establishes
	\eqref{eq:Eloglik}.
	
	Now note that for a positive constant $\Lambda$ the function $e^{y}-\Lambda y$
	is bounded below by both $-\Lambda y$ and $\Lambda y+2\Lambda\{1-\log2\Lambda\}$,
	hence $e^{y}-\Lambda y\geq\Lambda|y|+2\Lambda\min\{0,1-\log2\Lambda\}$.
	Since $\Lambda\min\{0,1-\log2\Lambda\}$ is non-increasing in $\Lambda$, \ref{asm1} implies that
	\[
	\begin{aligned}e^{F(t,x)}-\lambda(t,x)F(t,x) &
		\geq\min\left\{ e^{F(t,x)}-\Lambda_{L}F(t,x),e^{F(t,x)}-\Lambda_UF(t,x)\right\} \\
		& \geq\Lambda_{L}|F(t,x)|+2\Lambda_U\min\{0,1-\log(2\Lambda_U)\}.
	\end{aligned}
	\]
	Integrating both sides and using the norm equivalence relation~\eqref{eq:normequiv}
	shows that
	\[
	\begin{aligned}R(F) & \geq\Lambda_{L}\|F\|_{\mu,1}+2\Lambda_U\min\{0,1-\log(2\Lambda_U)\}\\
		& \geq\frac{2\Lambda_{L}}{\aconst}\|F\|_{\infty}+2\Lambda_U\min\{0,1-\log(2\Lambda_U)\}\\
		& \geq\frac{2\Lambda_{L}}{\aconst}\|F\|_{\mu,2}+2\Lambda_U\min\{0,1-\log(2\Lambda_U)\}.
	\end{aligned}
	\]
	The lower bound~\eqref{eq:coercive1} then follows from the second
	inequality. The last inequality shows that $R(F)$ is coercive on
	$(\ff,\left\langle \cdot,\cdot\right\rangle _{\mu})$. Moreover the
	same argument used to derive~\eqref{eq:emptaylor} shows that $R(F)$ is
	smooth and convex on $(\ff,\left\langle \cdot,\cdot\right\rangle
	_{\mu})$.  Therefore a unique minimizer $F^{*}$ of $R(F)$ exists in
	$(\ff,\left\langle \cdot,\cdot\right\rangle _{\mu})$.  Since \ref{asm2}
	implies there is a bijection between the equivalent classes of
	$(\ff,\left\langle \cdot,\cdot\right\rangle _{\mu})$ and the functions
	in $\ff$, $F^{*}$ is also the unique minimizer of $R(F)$ in
	$\ff$. Finally, since $e^{F(t,x)}-\lambda(t,x)F(t,x)$ is pointwise
	bounded below by $\lambda(t,x)\{1-\log\lambda(t,x)\}$,
	$R(F)\geq\int(\lambda-\lambda\log\lambda)d\mu=R(\log\lambda)$ for all
	$F\in\ff$.
\end{proof}

\subsection*{Proof of Lemma \ref{lem:stat-converge}}

\begin{proof}
	By a pointwise-measurable argument (Example 2.3.4 of~\citep{vaart}) it
	can be shown that all suprema quantities appearing below are
	sufficiently well behaved, so outer integration is not
	required. Define the Orlicz norm
	$\orlicz{X}=\inf\{C>0:\EE\Phi(|X|/C)\le 1\}$ where
	$\Phi(x)=e^{x^2}-1$.  Suppose the following holds:
	\EqArray
	\orlicz{\sup_{F\in\ff_\e^{\Psi_{n}}}\left|\{\Rhat_n(F)-\Rhat_n(0)\}-\{R(F)-R(0)\}\right|\,}
	&\leq&
	\kappa' J_{\ff_\e}/n^{1/4}, \label{eq:esup1}\\
	\orlicz{\sup_{G\in\ff_\e: \|G\|_\infty \leq 1}\Bigl|\|G\|_{\mun,1}-\|G\|_{\mu,1}\Bigr|\,}
	&\leq&
	\kappa''J_{\ff_\e}/n^{1/2},\label{eq:esup2}
	\EndEqArray
	where $J_{\ff_\e}$ is the complexity measure~\eqref{eq:entintegral},
	and $\kappa',\kappa''$ are universal constants. Then by Markov's
	inequality,~\eqref{eq:stat-converge} holds with probability
	at least $1-2\exp[-\{\eta n^{1/4}/(\kappa'J_{\ff_\e})\}^{2}]$, and
	\Eq
	\sup_{G\in\ff_\e: \|G\|_\infty \leq 1}\left\{
	\|G\|_{\mu,1}-\|G\|_{\mun,1}\right\} <1/\aconst
	\label{eq:psup}
\end{equation}
holds with probability at least
$1-2\exp[-\{n^{1/2}/(\aconst\kappa''J_{\ff_\e})\}^{2}]$.  Since
$\aconst>1$ and $\eta<1$, (\ref{eq:stat-converge}) and
(\ref{eq:psup}) jointly hold with probability at least
$1-4\exp[-\{\eta n^{1/4}/(\kappa\aconst J_{\ff_\e})\}^{2}]$.  The
lemma then follows if~\eqref{eq:psup}
implies~\eqref{eq:emp-normequiv}. Indeed, for any non-zero $F\in\ff$,
its normalization $G=F/\|F\|_{\infty}$ is in $\ff_\e$ by
construction~\eqref{eq:smallF}. Then~\eqref{eq:psup} implies that
$$
\frac{\|F\|_{\infty}}{\|F\|_{\mun,1}}=1/\|G\|_{\mun,1}\leq\aconst
$$
because
$$
1/\aconst>\|G\|_{\mu,1}-\|G\|_{\mun,1}\geq2/\aconst-\|G\|_{\mun,1},
$$
where the last inequality follows from the definition of $\aconst$~\eqref{eq:alphaconst}.

Thus it remains to establish~\eqref{eq:esup1} and~\eqref{eq:esup2},
which can be done by applying the symmetrization and maximal
inequality results in Sections 2.2 and 2.3.2 of~\cite{vaart}. Write
$\Rhat_n(F)=(1/n)\sum_{i=1}^{n}l_{i}(F)$ where
$l_{i}(F)=\int_{0}^{1}Y_i(t)e^{F(t,X_{i}(t))}dt-\Delta_{i}F(T_{i},X_{i}(T_{i}))$
are independent copies of the loss
\begin{equation}
	l(F)=\int_{0}^{1} Y(t)\cdot e^{F(t,X(t))}dt-\Delta\cdot F(T,X(T)),
	\label{eq:lossprocess}
\end{equation}
which is a stochastic process indexed by $F\in\ff$. As was shown in
Proposition~\ref{prop:expectedR}, $\EE\{l(F)\}=R(F)$. Let
$\zeta_{1},\cdots,\zeta_{N}$ be independent Rademacher random
variables that are independent of $Z=\{(X_i(\cdot),Y_i(\cdot),T_i)\}_{i=1}^{n}$. It
follows from the symmetrization Lemma 2.3.6 of~\cite{vaart} for
stochastic processes that the left hand side of~\eqref{eq:esup1} is
bounded by twice the Orlicz norm of
\EqArray\label{eq:mainsupbound}
&&\hskip-30pt
\sup_{F\in\ff_\e^{\Psi_{n}}}\left|\frac{1}{n}\sum_{i=1}^{n}\zeta_{i}\{l_{i}(F)-l_{i}(0)\}\right|
\nonumber\\
&& \leq
\frac{1}{n}\sup_{F\in\ff_\e^{\Psi_{n}}}\left|\sum_{i=1}^{n}\zeta_{i}\int_{0}^{1}
Y_i(t) \left\{ e^{F(t,X_{i}(t))}-1\right\} dt\right| \\
&&\quad + \frac{1}{n}\sup_{F\in\ff_\e^{\Psi_{n}}}\left|\sum_{i=1}^{n}\zeta_{i}\Delta_{i}F(T_{i},X_{i}(T_{i}))\right|. \nonumber
\EndEqArray

Now hold $Z$ fixed so that only $\zeta_{1},\cdots,\zeta_{n}$ are
stochastic, in which case the sum in the second line
of~\eqref{eq:mainsupbound} becomes a separable subgaussian
process. Since the Orlicz norm of $\sum_{i=1}^{n}\zeta_{i}a_{i}$ is
bounded by $(6\sum_{i=1}^{n}a_{i}^{2})^{1/2}$ for any constant
$a_{i}$, we obtain the following the Lipschitz property for any
$F_1,F_2\in\ff_\e^{\Psi_{n}}$:
\Array
&&\hskip-30pt
\orliczSqCond{\sum_{i=1}^{n}\zeta_{i}\int_{0}^{1}Y_i(t)\left\{ e^{F_1(t,X_{i}(t))}-e^{F_2(t,X_{i}(t))}\right\} dt}{\zeta|Z}\\
&&\leq 6\sum_{i=1}^{n}\left[\int_{0}^{1} Y_i(t)\left\{ e^{F_1(t,X_{i}(t))}-e^{F_2(t,X_{i}(t))}\right\} dt\right]^{2}\\
&&\leq 6e^{2\Psi_{n}}\sum_{i=1}^{n}\left(\int_{0}^{1} Y_i(t)\cdot |F_1(t,X_{i}(t))-F_2(t,X_{i}(t))|dt\right)^{2}\\
&&\leq 6e^{2\Psi_{n}}\sum_{i=1}^{n}\int_{0}^{1}Y_i(t)\{F_1(t,X_{i}(t))-F_2(t,X_{i}(t))\}^{2}dt\\
&&=    6ne^{2\Psi_{n}}\|F_1-F_2\|_{\mun,2}^{2},
\EndArray
where the second inequality follows from $|e^{x}-e^{y}|\leq e^{\max(x,y)}|x-y|$
and the last from the Cauchy-Schwarz inequality. Putting the Lipschitz
constant $(6n)^{1/2}e^{\Psi_{n}}$ obtained above into Theorem 2.2.4
of \cite{vaart} yields the following maximal inequality: There is
a universal constant $\kappa'$ such that
\Array
&&\hskip-30pt
\orliczCond{\sup_{F\in\ff_\e^{\Psi_{n}}}\left|\sum_{i=1}^{n}\zeta_{i}\int_{0}^{1}Y_i(t)
	\left\{ e^{F(t,X_{i}(t))}-1\right\} dt\right|\,}{\zeta|Z}\\
&&\leq \kappa' n^{1/2}e^{\Psi_{n}}\int_{0}^{\Psi_{n}}\left\{ \log\mathcal{N}(u,\ff_\e^{\Psi_{n}},\mun)\right\} ^{1/2}du\\
&&\leq \kappa' n^{1/2}e^{\Psi_{n}}\Psi_{n}J_{\ff_\e},
\EndArray
where the last line follows from~\eqref{eq:entintegral}. Likewise the
conditional Orlicz norm for the supremum of
$\left|\sum_{i=1}^{n}\zeta_{i}\Delta_{i}F(T_{i},X_{i}(T_{i}))\right|$
is bounded by $\kappa' J_{\ff_\e}n^{1/2}\Psi_{n}$.  Since neither
bounds depend on $Z$, plugging back into~\eqref{eq:mainsupbound}
establishes~\eqref{eq:esup1}:
\Array
&&\hskip-30pt
\orlicz{\sup_{F\in\ff_\e^{\Psi_{n}}}\left|\{\Rhat_n(F)-\Rhat_n(0)\}-\{R(F)-R(0)\}\right|\,}\\
&&\leq 2\kappa' J_{\ff_\e}\frac{\Psi_{n}e^{\Psi_{n}}}{n^{1/2}}
\left\{1+e^{-\Psi_{n}}\right\} \\
&& \leq4\kappa' \frac{J_{\ff_\e}}{n^{1/4}},
\EndArray
where $\Psi_{n}e^{\Psi_{n}}=n^{1/4}$ by~\eqref{eq:stopping}.  On noting that
$$
\|G\|_{\mun,1} =
\frac{1}{n}\sum_{i=1}^{n}\int_{0}^{1}Y_i(t)|G(t,X_{i}(t))|dt,\hskip10pt
\|G\|_{\mu,1}=\EE\left\{ \int_{0}^{1}Y(t)|G(t,X(t))|\,dt\right\},
$$
\eqref{eq:esup2} can be established using the same approach.
\end{proof}

\subsection*{Proof of Lemma~\ref{lem:emp-converge}}

\begin{proof}
	For $m<\mhat$, applying~\eqref{eq:emptaylor} to
	$\Rhat_n(\Fhat_{m+1})=\Rhat_n(\Fhat_m-\frac{\nu_{n}}{m+1}g_{\Fhat_m}^{\e})$
	yields
	\begin{equation}\label{eq:taylorbound}
		\begin{aligned}
			\Rhat_{n}(\Fhat_{m+1})
			&=
			\Rhat_{n}(\Fhat_{m})-\frac{\nu_{n}}{m+1}\left\langle
			g_{\Fhat_{m}},g_{\Fhat_{m}}^{\e}\right\rangle_{\mun}\\
			&\quad +\frac{\nu_{n}^{2}}{2(m+1)^{2}}\int(g_{\Fhat_{m}}^{\e})^{2}
			\exp\left\{ \Fhat_{m}-\rho (\Fhat_{m+1} - \Fhat_m)   \right\} d\mun\\
			&<\Rhat_{n}(\Fhat_{m})-\frac{\e\nu_{n}}{m+1}\|g_{\Fhat_{m}}\|_{\mun,2}+\frac{\nu_{n}^{2}e^{\Psi_{n}}}{2(m+1)^{2}},
		\end{aligned}
	\end{equation}
	where the bound for the second term is due to~\eqref{eq:epsgradient.alg}
	and the bound for the integral follows from
	$\int(g_{\Fhat_{m}}^{\e})^{2}d\mun=1$
	(Definition~\ref{def:eps.gradient} of an $\e$-gradient) and
	$\|\Fhat_{m}\|_{\infty}, \|\Fhat_{m+1}\|_{\infty}<\Psi_n$ for
	$m<\mhat$ (lines 5-6 of Algorithm~\ref{alg:haz}). Hence for
	$m\leq\mhat$, \eqref{eq:taylorbound} implies that
	$$
	\Rhat_{n}(\Fhat_{m})<\Rhat_{n}(0)
	+\sum_{m=0}^{\infty}\frac{\nu_{n}^{2}e^{\Psi_{n}}}{2(m+1)^{2}}<\Rhat_{n}(0)+1\leq2
	$$
	because $\nu_n^2 e^{\Psi_n} < 1$ under~\eqref{eq:shrinkfactor}. Since
	$\max_{m\leq\hat{m}}\|\Fhat_{m}\|_{\infty}<\Psi_{n}$, and using our
	assumption $\sup_{F\in\ff_{\e}^{\Psi_{n}}}|\Rhat_{n}(F)-R(F)|<1$ in
	the statement of the lemma, we have
	$$
	R(\Fhat_{m})
	\leq \Rhat_{n}(\Fhat_{m})+\left|\Rhat_{n}(\Fhat_{m})-R(\Fhat_{m})\right|
	<3.
	$$
	Clearly the minimizer $F^{*}$ also satisfies $R(F^{*})\leq R(0)<3$.
	Thus coercivity (\ref{eq:coercive2}) implies that
	$$
	\|\Fhat_{m}\|_{\infty},\|F^{*}\|_{\infty}<\alpha_{\ff}\beta_{\Lambda},
	\label{eq:F_bd}
	$$
	so the gap $\hat{\gamma}$ defined in~\eqref{eq:descentgap} is bounded
	as claimed.
	
	It remains to establish~\eqref{eq:emp-converge}, for
	which we need only consider the case
	$\Rhat_{n}(\Fhat_{\hat{m}})-\Rhat_{n}(F^{*})>0$.  The termination
	criterion $g_{\Fhat_{m}}=0$ in Algorithm~\ref{alg:haz} is never
	triggered under this scenario, because by Proposition~\ref{prop:empR} this
	would imply that $\Fhat_{\mhat}$ minimizes $\Rhat_{n}(F)$
	over the span of $\{\phihat_{nj}(t,x)\}_j$, which also contains $F^*$ (Remark \ref{rem:dim.equivalence}). Thus either
	$\mhat=\infty$, or the termination criterion $\| \Fhat_{\mhat}-\frac{\nu_{n}}{\mhat+1}\hat{g}_{\Fhat_{\mhat}}^{\e} \|_\infty \ge \Psi_n$ in line 5 of Algorithm \ref{alg:haz} is met. In the latter case
	\begin{equation}\label{eq:iter_lb}
		\begin{aligned}
			\Psi_{n}
			&\leq \left\Vert
			\Fhat_{\mhat}-\frac{\nu_{n}}{\mhat+1}g_{\Fhat_{\mhat}}^{\e}\right\Vert_{\infty}
			\leq\aconst\left\Vert \Fhat_{\mhat}-\frac{\nu_{n}}{\mhat+1}g_{\Fhat_{\mhat}}^{\e}\right\Vert _{\mun,2}\\
			&\leq \alpha_{\ff}\left(\sum_{m=0}^{\mhat-1}\frac{\nu_{n}}{m+1}+1\right)
	\end{aligned}\end{equation}
	where the inequalities follow from~\eqref{eq:emp-normequiv} and from
	$\|g_{\Fhat_m}^{\e}\|_{\mun,2}=1$. Since the sum is diverging, the inequality also holds for $\mhat$ sufficiently large (e.g. $\mhat=\infty$).
	
	Given that $F^{*}$ lies in the span
	of $\{\phihat_{nj}(t,x)\}_j$, the Taylor expansion~\eqref{eq:emptaylor} is valid for $\Rhat_n(F^{*})$.
	Since the remainder term in the expansion is non-negative, we have
	\Array
	\Rhat_n(F^{*})
	&=& \Rhat_n(\Fhat_m + F^{*} - \Fhat_m)\\
	&\ge& \Rhat_n(\Fhat_m)  + \left\langle
	g_{\Fhat_{m}},F^{*}-\Fhat_{m}\right\rangle_{\mun}.
	\EndArray
	Furthermore for $m \le \hat m$,
	\Array
	\left\langle g_{\Fhat_{m}},\Fhat_{m}-F^{*}\right\rangle_{\mun}
	&\leq& \|\Fhat_{m}-F^{*}\|_{\mun,2} \cdot \|g_{\Fhat_{m}}\|_{\mun,2}\\
	&\leq& \|\Fhat_{m}-F^{*}\|_{\infty} \cdot \|g_{\Fhat_{m}}\|_{\mun,2}\\
	&\leq& \hat{\gamma}\cdot \|g_{\Fhat_{m}}\|_{\mun,2}.
	\EndArray
	Putting both into~\eqref{eq:taylorbound} gives
	\Array
	\Rhat_{n}(\Fhat_{m+1})
	&<&
	\Rhat_{n}(\Fhat_{m})+\frac{\e\nu_{n}}{\hat{\gamma}(m+1)}
	\left\langle g_{\Fhat_{m}},F^{*}-\Fhat_{m}\right\rangle_{\mun}
	+\frac{\nu_{n}^{2}e^{\Psi_{n}}}{2(m+1)^{2}}\\
	&\leq&
	\Rhat_{n}(\Fhat_{m})+\frac{\e\nu_{n}}{\hat{\gamma}(m+1)}
	\{\Rhat_{n}(F^{*})-\Rhat_{n}(\Fhat_{m})\}+\frac{\nu_{n}^{2}e^{\Psi_{n}}}{2(m+1)^{2}}.
	\EndArray
	Subtracting $\Rhat_{n}(F^{*})$ from both sides above and denoting
	$\delta_{m}=\Rhat_{n}(\Fhat_{m})-\Rhat_{n}(F^{*})$,
	we obtain
	$$
	\delta_{m+1}
	<\left(1-\frac{\e\nu_{n}}{\hat{\gamma}(m+1)}\right)\delta_{m}
	+\frac{\nu_{n}^{2}e^{\Psi_{n}}}{2(m+1)^{2}}.
	$$
	Since the term inside the first parenthesis is between 0 and 1, solving
	the recurrence yields
	\Array
	\delta_{\mhat}
	&<&
	\delta_{0}\prod_{m=0}^{\mhat-1}
	\left(1-\frac{\e\nu_{n}}{\hat{\gamma}(m+1)}\right)
	+\nu_{n}^{2}e^{\Psi_{n}}\sum_{m=0}^{\infty}\frac{1}{2(m+1)^{2}}\\
	&\leq&
	\max\{0,\delta_{0}\}
	\exp\left(-\frac{\e}{\hat{\gamma}}\sum_{m=0}^{\mhat-1}\frac{\nu_{n}}{m+1}\right)+\nu_{n}^{2}e^{\Psi_{n}}\\
	&\leq&
	e\max\{0,\delta_{0}\}\exp\left(-\frac{\e}{\aconst\hat{\gamma}}\Psi_{n}\right)+\nu_{n}^{2}e^{\Psi_{n}},
	\EndArray
	where in the second inequality we used the fact that $0\leq1+y\leq e^{y}$
	for $|y|<1$, and the last line follows from~\eqref{eq:iter_lb}.
	
	The Lambert function~\eqref{eq:stopping} in $\Psi_{n}=W(n^{1/4})$ is
	asymptotically $\log y-\log\log y$, and in fact by Theorem 2.1
	of~\citep{hoorfar}, $W(y)\geq\log y-\log\log y$ for $y\geq e$. Since
	by assumption $n\geq55>e^{4}$, the above becomes
	$$
	\delta_{\mhat}
	<e\max\{0,\delta_{0}\}
	\left(\frac{\log  n}{4n^{1/4}}\right)^{\e/(\aconst\hat{\gamma})}+\nu_{n}^{2}e^{\Psi_{n}}.
	$$
	The last step is to control $\delta_{0}$, which is bounded
	by $1-\Rhat_{n}(F^{*})$ because $\Rhat_{n}(\Fhat_{0})=\Rhat_{n}(0)\leq1$.
	Then under the hypothesis $|\Rhat_{n}(F^{*})-R(F^{*})|<1$,
	we have
	$$
	\delta_{0}\leq 1-R(F^{*})+1<2-R(F^{*}).
	$$
	Since~\eqref{eq:coercive1}
	implies $R(F^{*})\geq2\Lambda_{U}\min\{0,1-\log(2\Lambda_{U})\}$,
	$$
	\delta_{0}
	<2-R(F^{*})\leq2
	+2\Lambda_{U}\max\{0,\log(2\Lambda_{U})-1\}<2\beta_{\Lambda}.
	$$
\end{proof}

\subsection*{Proof of Proposition~\ref{prop:consistency}}

\begin{proof}
	
	Let $\delta=\log n/(4n^{1/4})$ which is less than one for $n\geq 55 >
	e^{4}$.  Since $\aconst,\hat{\gamma}\geq1$ it follows that
	\begin{equation}
		\delta<\left(\frac{\log n}{4n^{1/4}}\right)^{\e/(\aconst\hat{\gamma})}.
		\label{eq:epsbound}
	\end{equation}
	Now define the following probability sets
	\Array
	S_{1}
	&=&
	\left\{ \sup_{F\in\ff_{\e}^{\Psi_{n}}}\left|\{\Rhat_{n}(F)-\Rhat_{n}(0)\}-\{R(F)-R(0)\}\right|<\delta/3\right\} \\
	S_{2}
	&=&\left\{ \left|\Rhat_{n}(0)-R(0)\right|<\delta/3\right\} \\
	S_{3}
	&=&\left\{ \left|\Rhat_{n}(F^{*})-R(F^{*})\right|<\delta/3\right\} \\
	S_{4}
	&=&\left\{ \mbox{\eqref{eq:emp-normequiv} holds}\right\} ,
	\EndArray
	and fix a sample realization from $\cap_{k=1}^{4}S_{k}$. Then the
	conditions required in Lemma \ref{lem:emp-converge} are satisfied with
	$\sup_{F\in\ff_{\e}^{\Psi_{n}}}|\Rhat_{n}(F)-R(F)|<2\delta/3$, so
	$\hat{\gamma}$ (and hence $\|\lboosthat\|_\infty$) is bounded and~\eqref{eq:emp-converge} holds. Since Algorithm~\ref{alg:haz} ensures that $\|\Fhat_{\mhat}\|_{\infty}<\Psi_{n}$, we have $\Fhat_{\mhat}\in \ff_{\e}^{\Psi_{n}}$ and therefore it also follows that $|\Rhat_{n}(\Fhat_{\mhat})-R(\Fhat_{\mhat})|<2\delta/3$. Combining~\eqref{eq:decomp1} and~\eqref{eq:decomp2} gives
	\Array
	\left\Vert \Fhat_{\mhat}-F^{*}\right\Vert _{\mu,2}^{2}
	&\leq&
	\frac{2}{\min_{t,x}(\lambda^{*}\wedge\lboosthat)}
	\left(\frac{2\delta}{3}+\frac{\delta}{3}
	+ \{\Rhat_{n}(\Fhat_{\mhat})-\Rhat_{n}(F^{*})\}\right)\\
	&<&
	\frac{2}{\min_{t,x}(\lambda^{*}\wedge\lboosthat)}
	\left(\delta+2e\beta_{\Lambda}\left(\frac{\log n}{4n^{1/4}}\right)^{\e/(\aconst\hat{\gamma})}
	+\frac{1}{16}\cdot\frac{\log n}{4n^{1/4}}\right)\\
	&<&
	\frac{13\beta_{\Lambda}}{\min_{t,x}(\lambda^{*}\wedge\lboosthat)}
	\left(\frac{\log n}{4n^{1/4}}\right)^{\e/(\aconst\hat{\gamma})},
	\EndArray
	where the second inequality follows from~\eqref{eq:emp-converge} and
	$\nu_{n}^{2} e^{\Psi_n}=\log n/(64n^{1/4})$, and the last from~\eqref{eq:epsbound}. Now, using the inequality $|e^{x}-e^{y}|\leq\max(e^{x},e^{y})|x-y|$ yields
	$$
	\left\Vert\lboosthat-\l^*\right\Vert _{\mu,2}^{2}
	<13\beta_{\L}\frac{\max_{t,x}(\l^*\vee\lboosthat)^2}{\min_{t,x}(\l^*\wedge\lboosthat)}
	\left(\frac{\log n}{4n^{1/4}}\right)^{\e/(\aconst\hat{\gamma})},
	$$
	and the stated bound follows from $F^*=\log\l$ since $\ff$ is correctly specified (Proposition~\ref{prop:expectedR}).

	The next task is to lower bound $\PP(\cap_{k=1}^{4}S_{k})$.
	It follows from Lemma \ref{lem:stat-converge} that
	$$
	\PP(S_{1}\cap S_{4})
	\geq
	1-4\exp\left\{ -\left(\frac{\log  n}
	{12\kappa\aconst J_{\ff_{\e}}}\right)^{2}\right\} .
	$$
	Bounds on $\PP(S_{2})$ and $\PP(S_{3})$ can be obtained using
	Hoeffding's inequality. Note from~\eqref{eq:loglik} that
	$\Rhat_{n}(0)=\sum_{i=1}^{n}\int_0^{1} Y_i(t)dt/n$ and
	$\Rhat_{n}(F^{*})=\sum_{i=1}^{n}l_{i}(F^{*})/n$ for the loss
	$l(\cdot)$ defined in~\eqref{eq:lossprocess}. Since $0\leq\int_0^{1}
	Y_i(t)dt\leq1$ and
	$-\|F^{*}\|_{\infty}<l(F^{*})\leq\|e^{F^{*}}\|_{\infty}+\|F^{*}\|_{\infty}$,
	$$
	\PP(S_{2})
	\geq
	1-2\exp\left\{ -2n^{1/2}\left(\frac{\log n}{12}\right)^{2}\right\},\
	\PP(S_{3})
	\geq
	1-2\exp\left\{ -2n^{1/2}\left(\frac{\log n}{36e^{\|F^{*}\|_{\infty}}}\right)^{2}\right\} .
	$$
	By increasing the value of $\kappa$ and/or replacing $J_{\ff_{\e}}$
	with $\max(1,J_{\ff_{\e}})$ if necessary, we can combine the
	inequalities to get a crude but compact bound:
	\begin{equation}\label{eq:crudebound}
		\PP\{\cap_{k=1}^{4}S_{k}\}
		\geq
		1-8\exp\left\{ -\left(\frac{\log  n}
		{\kappa\aconst J_{\ff_{\e}}e^{\|F^{*}\|_{\infty}}}\right)^{2}\right\}.
	\end{equation}
	Finally, since $\|F^*\|_\infty = \|\log\lambda\|_\infty < \max\{|\log\Lambda_L|,|\log\Lambda_U|\}$, we can replace $e^{\|F^*\|_\infty}$ in the probability bound above by $\Lambda_L^{-1}\vee\Lambda_U$.
\end{proof}

\subsection*{Proof of Proposition~\ref{prop:oracleineq}}

\begin{proof}
	It follows from \eqref{eq:L2projection} that $\l^*$ is the orthogonal projection of
	$\lambda$ onto $(\ff,\left\langle \cdot,\cdot\right\rangle _{\mu})$.
	Hence
	\Array
	\left\Vert \lboosthat-\lambda\right\Vert _{\mu,2}^{2}
	& = &
	\left\Vert e^{F^*}-\lambda\right\Vert _{\mu,2}^{2}
	+\left\Vert e^{\Fhat_{\mhat}}-e^{F^{*}}\right\Vert _{\mu,2}^{2}\\
	& = &
	\min_{F\in\ff}\left\Vert e^{F}-\lambda\right\Vert _{\mu,2}^{2}
	+\left\Vert e^{\Fhat_{\mhat}}-e^{F^{*}}\right\Vert _{\mu,2}^{2}\\
	&\leq&
	\min_{F\in\ff}\left\Vert e^{F}-\lambda\right\Vert _{\mu,2}^{2}
	+\max_{t,x}(\lambda^{*}\vee\lboosthat)^{2}\left\Vert \Fhat_{\mhat}-F^{*}\right\Vert _{\mu,2}^{2},
	\EndArray
	where the inequality follows from $|e^{x}-e^{y}|\leq\max(e^{x},e^{y})|x-y|$.
	Bounding the last term in the same way as Proposition~\ref{prop:consistency}
	completes the proof. To replace $e^{\|F^*\|_\infty}$ in \eqref{eq:crudebound} by $\Lambda_L^{-1}\vee\Lambda_U$, it suffices to show that $\Lambda_L \leq \lambda^*(t,x) \leq \Lambda_U$. Since the value of $\lambda^*$ over one of its piecewise constant regions $B$ is $\int_B\lambda d\mu/\mu(B)$, the desired bound follows from \ref{asm1}. We can also replace $\max_{t,x}(\l^*\vee\lboosthat)$ and $\min_{t,x}(\l^*\wedge\lboosthat)$ with $\max_{t,x}(\Lambda_U\vee\lboosthat)$ and $\min_{t,x}(\Lambda_L\wedge\lboosthat)$ respectively.
\end{proof}

\newpage
\centerline{\HugeCaps REFERENCES}
\bibliographystyle{abbrvnat}

\begingroup
\renewcommand{\section}[2]{}%
\bibliography{ref}
\endgroup

\end{document}